\documentclass[utf8]{frontiersSCNS} % for Science, Engineering and Humanities and Social Sciences articles
\usepackage{url,hyperref,lineno,microtype,subcaption}
\usepackage[onehalfspacing]{setspace}

\usepackage{multirow} % for table

\usepackage{pifont}
\newcommand{\cmark}{\ding{51}}
\newcommand{\xmark}{\ding{55}}
\def\keyFont{\fontsize{8}{11}\helveticabold }
\def\firstAuthorLast{Rakhimberdina {et~al.}} %use et al only if is more than 1 author
\def\Authors{Zarina Rakhimberdina\,$^{1,3}$, Quentin Jodelet\,$^{1,3}$, Xin Liu\,$^{2,3,*}$, Tsuyoshi Murata\,$^{1,3}$}
\def\Address{$^{1}$Department of Computer Science,
Tokyo Institute of Technology, Tokyo 152-8552, Japan \\
$^{2}$Artificial Intelligence Research Center, National Institute of Advanced Industrial Science and Technology, Tokyo 135-0064, Japan\\
$^{3}$AIST-Tokyo Tech Real World Big-Data Computation Open Innovation Laboratory, Tokyo 152-8550, Japan}
\def\corrAuthor{Xin Liu}

\def\corrEmail{xin.liu@aist.go.jp}

\begin{document}
\onecolumn
\firstpage{1}

\title[Natural Image Reconstruction from fMRI using Deep Learning]{Natural Image Reconstruction from fMRI using Deep Learning: A Survey} 

\author[\firstAuthorLast ]{\Authors} %This field will be automatically populated
\address{} %This field will be automatically populated
\correspondance{} %This field will be automatically populated

%\extraAuth{}% If there are more than 1 corresponding author, comment this line and uncomment the next one.
\extraAuth{Zarina Rakhimberdina \\  zarina.rakhimberdina@net.c.titech.ac.jp}

\maketitle

\begin{abstract}
%%% Leave the Abstract empty if your article does not require one, please see the Summary Table for full details.

With the advent of brain imaging techniques and machine learning tools, much effort has been devoted to building computational models to capture the encoding of visual information in the human brain.
One of the most challenging brain decoding tasks is the accurate reconstruction of the perceived natural images from brain activities measured by functional magnetic resonance imaging (fMRI). 
%The progress in natural image reconstruction has many potential applications, including in brain-computer interfaces and neuroprosthetics. 
In this work, we survey the most recent deep learning methods for natural image reconstruction from fMRI. 
We examine these methods in terms of architectural design, benchmark datasets, and evaluation metrics and present a fair performance evaluation across standardized evaluation metrics. 
Finally, we discuss the strengths and limitations of existing studies and present potential future directions.

\tiny
 \keyFont{ \section{Keywords:} Natural Image Reconstruction, fMRI, Brain Decoding, Neural Decoding, Deep Learning} %All article types: you may provide up to 8 keywords; at least 5 are mandatory.
\end{abstract}

\section{Introduction}

\subsection{Visual decoding using fMRI}
Many brain imaging studies focus on decoding how the human brain represents information about the outer world. %, whether it is sensory stimuli, cognitive state, or motor information.
Considering that, the majority of external sensory information is processed by the human visual system \citep{logothetis_visual_1996}, a need for deeper understanding of visual information processing in the human brain %and the increasing amount of brain imaging data 
encourages building complex computational models that can characterize the content of visual stimuli. This problem is referred to as \textit{human visual decoding} of perceived images and has gained increasing attention.

A great advancement in recent neuroscience research has been achieved through functional magnetic resonance imaging (fMRI) \citep{poldrack_progress_2015, nestor_face_2020}. The fMRI technique captures neural activity in the brain by measuring variations in blood oxygen levels \citep{ogawa_brain_1990, bandettini_twenty_2012}. Among the various brain imaging techniques, fMRI is noninvasive and has a high spatial resolution. These characteristics allow fMRI to be used in a wide range of problems, including neurological disorder diagnosis \citep{zhang_survey_2020, rakhimberdina_population_2020} and human visual decoding \citep{haxby_distributed_2001, kamitani_decoding_2005, horikawa_generic_2017}. The recent progress in human visual decoding has shown that beyond merely encoding the information about visual stimuli \citep{poldrack_progress_2015}, brain activity captured by fMRI can be used to reconstruct visual stimuli information \citep{roelfsema_mind_2018, kay_identifying_2008}. 

\begin{figure*}
\centering
\includegraphics[width=12cm]{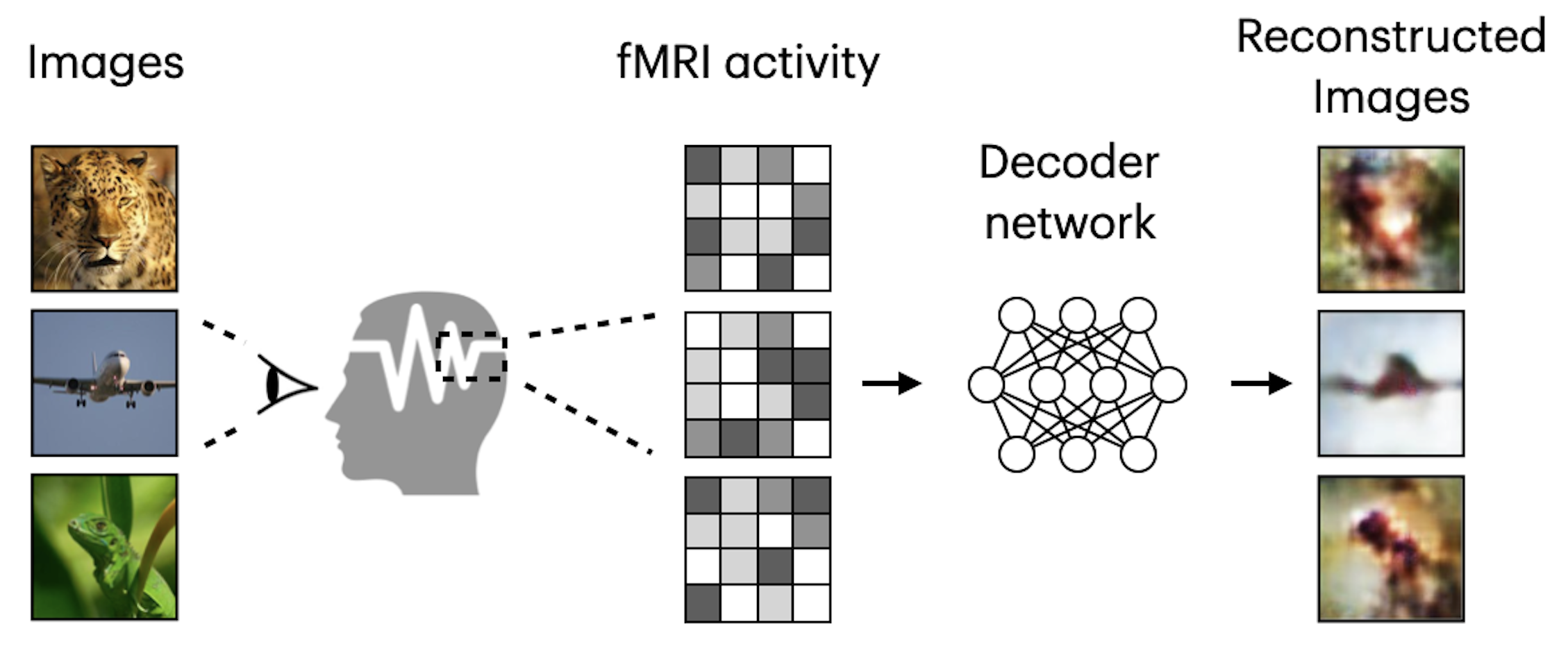}
\caption{Framework diagram for natural image reconstruction task.}
\label{fig:framework}
\end{figure*}

Based on the target task, human visual decoding can be categorized into stimuli category classification, stimuli identification, and reconstruction \citep{naselaris_encoding_2011}. 
In classification, brain activity is used to predict discrete object categories of the presented stimuli \citep{haxby_distributed_2001, horikawa_generic_2017}. 
The goal of identification is to identify a specific
stimulus corresponding to the given pattern of brain activity from a known set of stimuli images \citep{kay_identifying_2008, naselaris_encoding_2011}. In both identification and reconstruction, we aim to recover image-specific details, such as object position, size, and angle. %In \citep{kay_identifying_2008}, the authors have shown that patterns of fMRI activity "contain a considerableamount of stimulus-related information" to perform image identification suggesting the possibility of stimuli reconstruction from brain activity. 
However, reconstruction is a more challenging task, in which a replica of the stimulus image needs to be generated for a given fMRI signal (see Figure \ref{fig:framework}). Furthermore,  stimulus-related information encoded in the fMRI activity, which allows high-accuracy identification, may only partially characterize stimuli images and thus be insufficient for proper image reconstruction \citep{kay_identifying_2008, st-yves_generative_2018}. With the development of sophisticated image reconstruction methods and the increasing amount of brain imaging data, more attention has been directed toward visual stimuli reconstruction from fMRI activity in the visual cortex \citep{miyawaki_visual_2008, naselaris_bayesian_2009, van_gerven_neural_2010}. 
%Visual reconstruction is considered the most demanding  of visual decoding tasks because it requires accurate image reconstruction from limited and noisy  high-dimensional fMRI data. 
fMRI-based visual reconstruction can improve our understanding of the brain's visual processing mechanisms, and researchers can incorporate this knowledge into the development of brain--computer interfaces. 

\subsection{Natural image reconstruction}
The variety of visual stimuli used in visual reconstruction tasks can range from simple low-level detail images, such as Gabor wavelets and domino patterns \citep{thirion_inverse_2006}, to more elaborate images depicting alphabetical characters, digits \citep{van_gerven_neural_2010, schoenmakers_linear_2013}, natural objects, and scenes \citep{haxby_distributed_2001, horikawa_generic_2017}. The image reconstruction task for low-level detail stimuli does not require expressive models, and linear mapping is usually sufficient for learning effective reconstruction \citep{miyawaki_visual_2008}. Among the variety of visual stimuli, natural images are considered the most challenging, as they require accurate reconstruction of color, shape, and higher-level perceptual features. 

Similar to \cite{shen_deep_2019}, we refer to the task of visual stimuli reconstruction from fMRI as \textit{natural image reconstruction}, where stimuli are drawn from a database of natural images. 
%Let $X = \{x_1, x_2, ..., x_N\}$ and  $V = \{v_1, v_2, ..., v_N\}$ denote a set of stimulus images and a set of corresponding fMRI activity patterns extracted from the visual cortex, respectively.  The dataset comprises $N$ paired samples $S = \{(x_i, v_i)|x_i \in X, v_i \in V\}_{i=1}^N$.
The goal of neural decoding models is to learn a mapping function $f: \mathcal{V} \rightarrow \mathcal{X}$, where $\mathcal{X}$ and $\mathcal{V}$ denote two sets corresponding to stimulus images and fMRI activity patterns extracted from the visual cortex. A framework diagram for visual reconstruction is shown in Figure \ref{fig:framework}.

The main challenges of natural image reconstruction include the following. First, the reconstruction quality must be good enough to capture the similarity between reconstructed and original images on multiple levels. In contrast to low-resolution image stimuli, such as shape or character patterns, good-quality reconstruction of natural images requires that both  lower-level details and high-level semantic information be preserved. %Therefore, reconstructions from noisy fMRI activity can only partially characterize stimulus images at the current stage. 
Second, brain’s visual representations %in the human visual system 
are invariant to different objects or image details, which is essential for object recognition, but imply that brain activation patterns are not necessarily unique for a given stimulus object \citep{st-yves_generative_2018, quiroga_invariant_2005}.
Finally, the lack of a standardized evaluation procedure for assessing the reconstruction quality makes it difficult to compare the existing methods. %While the first two challenges can be a subject of ongoing research, 
In this work, we will primarily focus on the solution to the third challenge.

\textbf{Contributions}. 
The topic of natural image reconstruction from fMRI is relatively new and has attracted much interest over the last few years. The related surveys on the field of natural encoding and decoding of visual input give a broad overview of the existing techniques to extract information from the brain \citep{roelfsema_mind_2018, nestor_face_2020} and focus on the traditional machine learning methods \citep{chen_survey_2014}. %\citep{ben_yedder_deep_2020, min_deep_2017, zhang_review_2020}, 
To our knowledge, there is no comprehensive survey on the topic of natural image reconstruction from fMRI using deep learning. Given the lack of a standardized evaluation process in terms of the benchmark dataset and standard metrics, our main contribution is to provide the research community with a fair performance comparison for existing methods. 

In this survey, we provide an overview of the deep learning-based natural image reconstruction methods. We discuss the differences in architecture, learning paradigms, and advantages of deep learning models over traditional methods. %We provide a structured overview of the proposed research works arranged according to different criteria and discuss the advantages and disadvantages of each method. 
In addition, we review the evaluation metrics and compare models on the same benchmark: the same metrics and the same dataset parameters. The proposed standardised evaluation on a common set of metrics offers an opportunity to fairly evaluate and track new emerging methods in the field.

The rest of this paper is organized as follows. In Section \ref{sec:datasets} and Section \ref{sec:approaches}, we introduce popular publicly available datasets for natural image reconstruction and %In Section \ref{sec:DL}, we provide an overview of popular deep learning models, and 
review recent state-of-the-art deep learning models for natural image reconstruction, respectively. Then, we provide an overview of the evaluation metrics in Section \ref{sec:mertics}, and presents a fair comparative evaluation of the methods in Section \ref{sec:comp_eval}. Finally, we discuss the main challenges and possible future directions of this work. Section \ref{sec:conlusion} concludes the paper. %in Section \ref{sec:discussion} discusses the main challenges, current trends, and future directions of natural image reconstruction. Finally, Section \ref{sec:conlusion} concludes the paper.

\begin{table*}[h]
\caption{Characteristics of benchmark datasets.}
\label{table:datasets_description}
\resizebox{\linewidth}{!}{
\begin{tabular}{lllcccc}
\hline
\multicolumn{1}{c}{\textbf{Reference}} &
  \multicolumn{2}{c}{\textbf{Dataset}} &
  \textbf{\begin{tabular}[c]{@{}c@{}}Number of \\ Subjects\end{tabular}} &
  \textbf{\begin{tabular}[c]{@{}c@{}}Image stimuli \\ Train/Test\end{tabular}} &
  \textbf{\begin{tabular}[c]{@{}c@{}}Repetition time \\ Train/Test\end{tabular}} &
  \textbf{ROIs} \\ \hline
\cite{vanrullen_reconstructing_2019}   & \multicolumn{2}{l}{\textit{Faces}}                                         & 4 & 88/20    & n/a  & n/a                \\ \hline
\cite{kay_identifying_2008}            & \multicolumn{2}{l}{\textit{vim-1}}                                         & 2 & 1750/120 & 2/13 & V1, V2, V3, V4, LO \\ \hline
\cite{horikawa_generic_2017} &
  \multicolumn{2}{l}{\textit{Generic Object Decoding}} &
  5 &
  1200/50 &
  1/35 &
  \multirow{4}{*}{\begin{tabular}[c]{@{}c@{}}V1, V2, V3, V4, \\ LOC, FFA, PPA\end{tabular}} \\ \cline{1-6}
\multirow{3}{*}{\cite{shen_deep_2019}} & \multirow{3}{*}{\textit{Deep Image Reconstruction}} & Natural Images       & 3 & 1200/50  & 5/24 &                    \\ \cline{3-6}
                                       &                                                     & Artificial Shapes    & 3 & 0/40     & 0/20 &                    \\ \cline{3-6}
                                       &                                                     & Alphabetical Letters & 3 & 0/10     & 0/12 &                    \\ \hline
\end{tabular}
}
\end{table*}

\section{Benchmark datasets} \label{sec:datasets}
This section summarizes the publicly available benchmark datasets used in deep learning-based natural image reconstruction from fMRI activity. While there exist a variety of datasets used for stimuli reconstruction, such as binary contrast patterns (BCP) \citep{miyawaki_visual_2008}, \texttt{69} dataset of handwritten digits \citep{van_gerven_neural_2010}, \texttt{BRAINS} dataset of handwritten characters \citep{schoenmakers_linear_2013}, we focus on the datasets with higher level of perceptual complexity of presented stimuli: dataset of faces, grayscale natural images, and natural images from Imagenet.  Each sample of these datasets represents a labeled pair -- fMRI recording paired with the relevant stimuli image. Several distinctive characteristics of each dataset are presented in Table \ref{table:datasets_description}.

\begin{figure}
\begin{center}
\includegraphics[width=8cm]{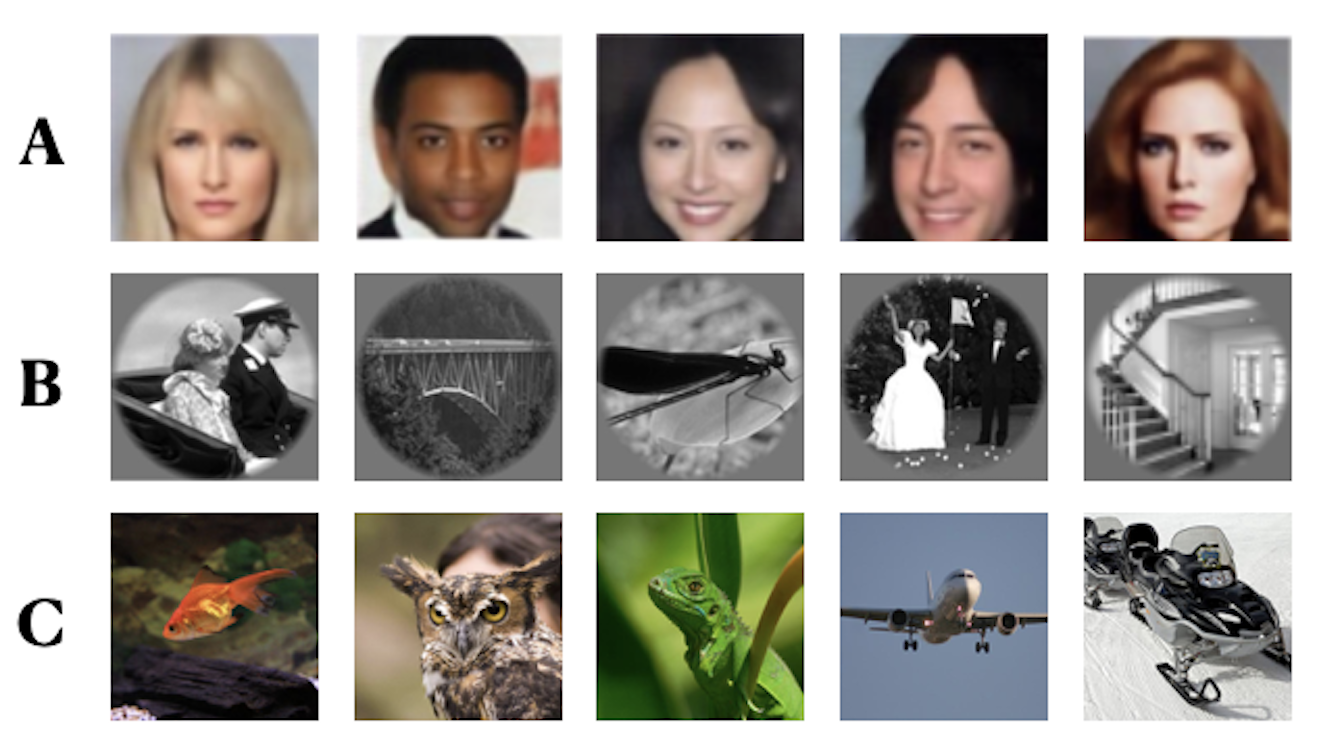}
\end{center}
\caption{Samples for natural stimuli: \textbf{(A)} images from Faces dataset \citep{vanrullen_reconstructing_2019}; \textbf{(B)} grayscale natural images from \texttt{vim-1} dataset \citep{kay_identifying_2008}; \textbf{(C)} natural images from \texttt{GOD} \citep{horikawa_generic_2017} and  \texttt{DIR} \citep{shen_deep_2019} datasets.}
\label{fig:stimuli}
\end{figure}

\textbf{Faces.} 
\cite{vanrullen_reconstructing_2019} used facial stimuli to reconstruct human faces from fMRI activity using deep neural networks\footnote{The fMRI dataset is available at \url{https://openneuro.org/datasets/ds001761}.}. The facial stimuli were drawn randomly from the CelebA  dataset \citep{liu_deep_2015}, and four healthy subjects participated in the experiment. The samples of stimuli images are shown in Figure \ref{fig:stimuli} A. 

\textbf{\texttt{vim-1} dataset of grayscale natural images} was acquired to study how natural images are represented by the human visual system\footnote{The dataset is available at \url{http://crcns.org/data-sets/vc/vim-1}. } \citep{kay_identifying_2008}.
The stimuli comprise a set of 1870 grayscale 500 $\times$ 500 pixels natural images of real-world objects, animals, and indoor and outdoor scenes (the samples are shown in Figure \ref{fig:stimuli} B).
Natural images were obtained from the Corel Stock Photo Libraries \citep{corel_corporation_corel_1994}, the Berkeley database of human segmented natural images\footnote{\url{https://www2.eecs.berkeley.edu/Research/Projects/CS/vision/grouping/segbench/}} \citep{martin_database_2001}, and an image collection from the authors.  %fMRI data comprises recordings from five regions of interest in visual cortices: V1, V2, V3, V4, and the lateral occipital area (LO).
Two healthy subjects with normal or corrected-to-normal vision were involved in the fMRI data acquisition. %For the training set, there are 1750 images, and each image was presented twice. For the test set, the total number of images is 120 presented 13 times each.

\textbf{Natural images from Imagenet. }
Two natural image datasets released by Kamitani Lab are widely used in image reconstruction. The first dataset, also known as the \texttt{Generic Object Decoding}\footnote{The dataset can be acquired from \url{http://brainliner.jp/data/brainliner/Generic\_Object\_Decoding}.} dataset or \texttt{GOD} for short, was originally used by \cite{horikawa_generic_2017} for the image classification task from the fMRI data and was later adopted for image reconstruction \citep{beliy_voxels_2019, ren_reconstructing_2021}. The dataset consists of pairs of high-resolution 500 $\times$ 500 pixels stimuli images (see Figure \ref{fig:stimuli} C)  and the corresponding fMRI recordings. fMRI scans were obtained from five healthy subjects; the stimuli images were selected from the ImageNet \linebreak dataset \citep{deng_imagenet_2009}  and span across 200 object categories.  %The training set comprises 1200 natural images from 150 object categories, and each image was presented once. The test dataset comprises 50 images from 50 object categories, and each fMRI recording was repeated 35 times for each stimulus image. 

The second dataset based on the natural image dataset was acquired for the image reconstruction task \citep{shen_deep_2019, shen_end--end_2019}. It is publicly available at  OpenNeuro\footnote{\url{https://openneuro.org/datasets/ds001506/versions/1.3.1}} , where it is cited  as \texttt{Deep Image Reconstruction}. We refer to this dataset as \texttt{Deep Image Reconstruction} or \texttt{DIR} for short. The \texttt{DIR} dataset contains 1,250 stimuli images that are identical to the ones used in \texttt{GOD}. Because of different image presentation experiments, in which training and test image stimuli were repeated 5 and 24 times respectively, the training set of the \texttt{DIR} dataset consists of a larger number of stimuli-fMRI pairs (5 $\times$ 1,200 samples) compared to the \texttt{GOD}. Three healthy subjects were involved in the image presentation. An appealing feature of this dataset is that, in addition to natural images, the dataset contains artificial shapes and alphabetical letters. % similar to \citep{miyawaki_visual_2008}, which were used only as test data in the original experiment by \citep{shen_deep_2019, shen_end--end_2019}. 
The artificial shapes dataset consists of 40 images -- a combination of eight colors and five geometric shapes. The alphabetical letters dataset consists of 10 letters (A, C, E, I, N, O, R, S, T, U) of consistent brightness and color.

\section{Deep learning-based approaches for natural image reconstruction} \label{sec:approaches} 

% traditional methods
Before deep learning, the traditional methods in natural image reconstruction estimated a linear mapping from fMRI signals to hand-crafted image features using linear regression models \citep{kay_identifying_2008, fujiwara_modular_2013, naselaris_bayesian_2009}. These methods primarily focus on extracting predefined low-level features from stimulus images, such as local image structures or features of Gabor filters \citep{fang_reconstructing_2020, beliy_voxels_2019}. %and, therefore, are not successful in reconstructing complex natural images 

% DNN in general and in visual decoding
In recent years, deep neural networks (DNNs) have significantly advanced computer vision research, replacing  models based on hand-crafted features. In particular, DNN models have achieved better accuracy and improved image quality in various computer vision tasks, including image classification \citep{krizhevsky_imagenet_2012}, image segmentation \citep{chen_semantic_2015}, and image restoration \citep{zhang_beyond_2017}. In visual decoding tasks using brain imaging data, deep learning approaches have been applied to image classification \citep{haxby_distributed_2001, horikawa_generic_2017}, object segmentation \citep{kamnitsas_efficient_2017}, and natural image reconstruction \citep{shen_deep_2019, shen_end--end_2019}. They were shown to be more powerful than traditional methods \citep{zhang_survey_2020, kriegeskorte_deep_2015} primarily due to the multilayer architecture allowing to learn nonlinear mappings from brain signals to stimulus images \citep{beliy_voxels_2019, shen_end--end_2019}.

% DNN in natural image reconstruction task
Motivated by the success of deep learning in image generation, many recent studies have widely used DNN models in natural image reconstruction for several reasons. First, the deep learning framework conforms to some degree to the visual encoding--decoding process occurring in the hierarchical regions of the human visual system \citep{pinto_high-throughput_2009, krizhevsky_imagenet_2012, schrimpf_brain-score_2018}. Second, the application of deep generative models allows the synthesis of high-quality natural-looking images, which is achieved by learning the underlying data distribution \citep{goodfellow_generative_2014}.
Additionally, the training process can be aided by models pretrained on larger image datasets 
\citep{shen_deep_2019, shen_end--end_2019}. 

In this section, we present the evolution of the state-of-the-art deep learning-based methods for natural image reconstruction. We analyze them in terms of %design strategy (two-stage or end-to-end), 
DNN architecture, use of pretraining, and the choice of the dataset. The most popular deep learning models used in natural image reconstruction tasks include non-generative methods such as convolutional neural networks, encoder--decoder-based frameworks \citep{kingma_auto-encoding_2014}; and generative methods, such as adversarial networks \citep{goodfellow_generative_2014} and variational autoencoders \citep{larsen_autoencoding_2016}. %We describe each of them in greater detail in the following sections.
%For easy reference, we give each method a name, the first part of which corresponds to the first author's name, and the second part captures the central idea of the framework. 
A comparison of the surveyed methods is presented in Table \ref{table:baselines_1}.

%\begin{landscape}
\begin{table*}[h]
\caption{Comparative table of the surveyed works. E2E represents end-to-end training. Loss denotes the loss function (MAE: mean absolute error; MSE: mean squared error; KL: KL divergence; Adv: adversarial loss; Cos: cosine similarity. The links to the source code are valid as of November, 2021.}
\label{table:baselines_1}
%\arrayrulecolor[rgb]{0.8,0.8,0.8}
\resizebox{\linewidth}{!}{
\begin{tabular}{llclcclc}
\hline
\multicolumn{1}{c}{\textbf{Method}} &
  \multicolumn{1}{c}{\textbf{Authors}} &
  \textbf{Year} &
  \textbf{Datasets} &
  \textbf{Loss} &
  \textbf{E2E} &
  \multicolumn{1}{c}{\textbf{Pre-training}} &
  \textbf{Public code} \\ \hline
SeeligerDCGAN &
  Seeliger et al.  &
  2018 &
  \begin{tabular}[c]{@{}l@{}}BRAINS\\ vim-1\\ GOD\end{tabular} &
  \begin{tabular}[c]{@{}c@{}}MAE\\ MSE\end{tabular} &
  no &
  \begin{tabular}[c]{@{}l@{}}Generator pre-trained on  ImageNet \\ \citep{chrabaszcz_downsampled_2017}, Microsoft COCO \\ \citep{lin_microsoft_2014}, datasets from \cite{maaten_new_2009} \\ and \cite{schomaker_forensic_2000}.  AlexNet-based \\ Comparator  trained on ImageNet.\end{tabular} &
  no \\ \hline
StYvesEBGAN &
  St-Yves and Naselaris  &
  2018 &
  vim-1 &
  \begin{tabular}[c]{@{}c@{}}MSE\\ Adv\end{tabular} &
  no &
  \begin{tabular}[c]{@{}l@{}}The denoiser and generator were pretrained \\ on 32   $\times$ 32 color images from the CIFAR-10 \\ dataset  \citep{krizhevsky_learning_2009} \end{tabular} &
  \href{https://github.com/styvesg/gan-decoding-supplementary}{yes} \\ \hline
ShenDNN(+DGN) &
  Shen et al. &
  2019 &
  \begin{tabular}[c]{@{}l@{}}DIR:\\ Natural images,\\ Artificial Shapes, \\ Alphabetical Letters\end{tabular} &
  MSE &
  no &
  \begin{tabular}[c]{@{}l@{}}VGG-19 pre-trained on ImageNet. \\ Pre-trained DGN \citep{dosovitskiy_inverting_2016}.\end{tabular} &
  \href{https://github.com/KamitaniLab/DeepImageReconstruction}{yes} \\ \hline
ShenGAN &
  Shen et al. &
  2019 &
  \begin{tabular}[c]{@{}l@{}}DIR:\\ Natural images,\\ Artificial Shapes, \\ Alphabetical Letters\end{tabular} &
  \begin{tabular}[c]{@{}c@{}}MSE\\ Adv\end{tabular} &
  yes &
  \begin{tabular}[c]{@{}l@{}}Caffenet-based Comparator  pre-trained \\ on ImageNet\end{tabular} &
  \href{https://github.com/KamitaniLab/End2EndDeepImageReconstruction}{yes} \\ \hline
BeliyEncDec &
  Beliy et al. &
  2019 &
  \begin{tabular}[c]{@{}l@{}}GOD \\ vim-1\end{tabular} &
  \begin{tabular}[c]{@{}c@{}}MSE\\ Cos\\ MAE\end{tabular} &
  no &
  \begin{tabular}[c]{@{}l@{}}Pretrained AlexNet-based encoder\end{tabular} &
  \href{https://github.com/WeizmannVision/ssfmri2im}{yes} \\ \hline
VanRullenVAE-GAN &
  VanRullen and Reddy &
  2019 &
  Faces &
  \begin{tabular}[c]{@{}c@{}}MSE\\ Adv\end{tabular} &
  no &
  Pre-trained on CelebA dataset &
  \href{https://github.com/rufinv/VAE-GAN-celebA}{yes} \\ \hline
GazivEncDec &
  Gaziv et al. &
  2020 &
  \begin{tabular}[c]{@{}l@{}}GOD \\ vim-1\end{tabular} &
  \begin{tabular}[c]{@{}c@{}}MSE\\ Cos\\ MAE\end{tabular} &
  no  &
  \begin{tabular}[c]{@{}l@{}}Pretrained AlexNet-based encoder \end{tabular} &
  no \\ \hline
QiaoGAN-BVRM &
  Qiao et al.  &
  2020 &
  vim-1 &
  MSE &
  no &
  \begin{tabular}[c]{@{}l@{}}Generator of BigGAN pre-trained \\ on ImageNet\end{tabular} &
  \href{https://github.com/KaiQiao1992/ETECRM}{partially} \\ \hline
FangSSGAN &
  Fang et al.  &
  2020 &
  \begin{tabular}[c]{@{}l@{}}DIR:\\ Natural images\end{tabular} &
  \begin{tabular}[c]{@{}c@{}}MAE\\ Adv\end{tabular} &
  no &
  - &
  \href{https://github.com/duolala1/Reconstructing-Perceptive-Images-from-Brain-Activity-by-Shape-Semantic-GAN}{partially} \\ \hline
MozafariBigBiGAN &
  Mozafari et al.  &
  2020 &
  GOD &
  Adv &
  no &
  BigBiGAN pre-trained on ImageNet &
  no \\ \hline
RenD-VAE/GAN &
  Ren et al.  &
  2021 &
  \begin{tabular}[c]{@{}l@{}}BCP\\ 6-9\\ BRAINS\\ GOD\end{tabular} &
  \begin{tabular}[c]{@{}c@{}}KL\\ Adv\end{tabular} &
  no &
  \begin{tabular}[c]{@{}l@{}}Model pre-trained on external \\ data from ImageNet\end{tabular} &
  no \\ \hline
\end{tabular}
}
%\arrayrulecolor{black}
\end{table*}
%\end{landscape}

%\subsection{Convolutional neural network} \label{sec:cnn}
\subsection{Non-generative methods} \label{sec:cnn}
\textbf{Convolutional neural network (CNN).}
Compared to a simpler multilayer feed-forward neural network, which disregards the structural information of input images, the CNN has a better feature extraction capability because of the information filtering performed by convolutional layers within a neighborhood of pixels \citep{lecun_backpropagation_1989}. Stacking convolutional layers on top of each other allows learning hierarchical visual features of input images, known as feature abstraction. The lower CNN layers learn low-level details, whereas the higher CNN layers extract global high-level visual information from images  \citep{mahendran_understanding_2015}. The use of CNNs is ubiquitous in image processing tasks, including image reconstruction. Specifically, encoder--decoder \citep{beliy_voxels_2019, gaziv_self-supervised_2020}, U-Net \citep{fang_reconstructing_2020}, generative adversarial network \citep{goodfellow_generative_2014}, and variational autoencoder \citep{kingma_auto-encoding_2014} are popular architectures that adopt stacked convolutional layers to extract features at multiple levels.  

\begin{figure}[h]
\begin{center}
\includegraphics[width=8cm]{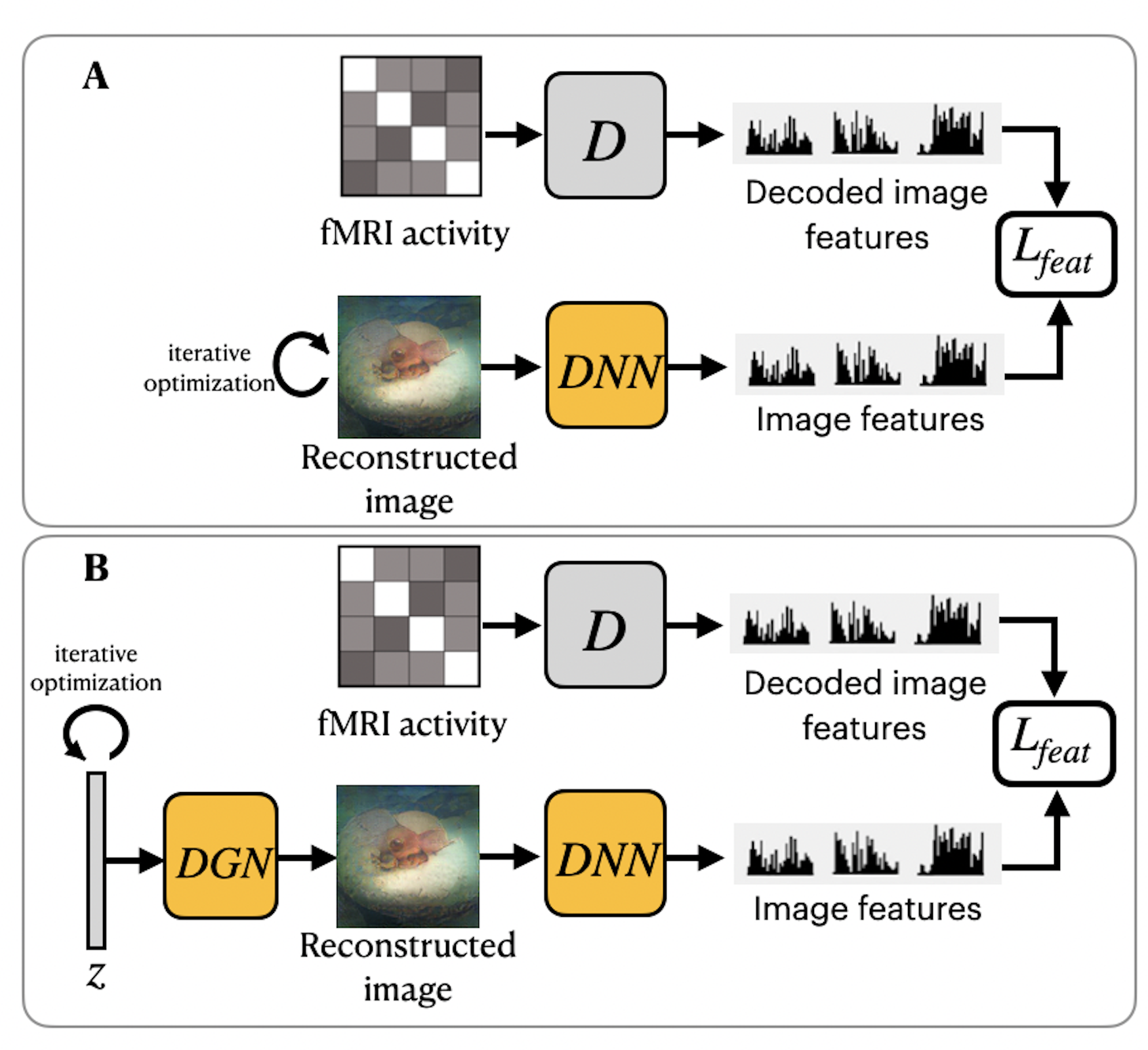}
\end{center}
\caption{Overview of two variations of frameworks proposed by \cite{shen_deep_2019}: \textbf{(A)} \texttt{ShenDNN} and \textbf{(B)} \texttt{ShenDNN+DGN}. The yellow color denotes the use of pretrained components.%: VGG-19 (DNN) available at \url{https://github.com/BVLC/caffe/wiki/Model-Zoo} and DGN by \citep{dosovitskiy_inverting_2016} available at \url{https://github.com/dosovits/caffe-fr-chairs}.
}
\label{fig:shen2019a}
\end{figure}

%\textbf{\texttt{ShenDNN}}. 
Shen et al. \citep{shen_deep_2019} utilized a pretrained VGG-19-based DNN to extract hierarchical features from stimuli images (see Figure \ref{fig:shen2019a} A). The DNN consists of sixteen convolutional layers followed by three fully connected layers. This method was motivated by the finding that hierarchical image representations obtained from different layers of deep neural network correlate with brain activity in the visual cortex \citep{eickenberg_seeing_2017, horikawa_generic_2017}. Using this fact, one can establish a hierarchical mapping from fMRI signals in the low/high-level areas of visual cortices to the corresponding low/high-level features from the DNN.
For this task, the authors implemented a feature decoder ${D}$ that maps fMRI activity patterns to multilayer DNN features. The decoder ${D}$ is trained on the train set before the reconstruction task, using the method from \cite{horikawa_generic_2017}. % $Z_{V}^{l} = \boldsymbol{D}(V)$, where $l$ represents the DNN layer.
These decoded fMRI features correspond to the hierarchical image features obtained from DNN. % $Z_{X}^{l} = \boldsymbol{DNN}(X)$.
The optimization is performed on the feature space by minimizing the difference between the hierarchical DNN features of the image and multilayer features decoded from fMRI activity. %The pretrained VGG-19 model is available from \url{https://github.com/BVLC/caffe/wiki/Model-Zoo}.

\textbf{Deterministic encoder--decoder models. } \label{sec:enc-dec}
In deep learning, %encoder--decoder models are multilayer models that incorporate convolutional layers. 
encoder--decoder models are widely used in image-to-image translation \citep{isola_image--image_2017} and sequence-to-sequence models \citep{cho_learning_2014}. They learn the mapping from an input domain to an output domain via a two-stage architecture: an encoder ${E}$ that compresses the input vector $\mathbf{x}$ to the latent representation $\mathbf{z} = {E}(\mathbf{x})$ and a decoder $\mathbf{y}={D}(\mathbf{z})$ that produces the output  vector $\mathbf{y}$ from the latent representation $\mathbf{z}$ \citep{minaee_image_2021}. The compressed latent representation vector $\mathbf{z}$ serves as a bottleneck, which encodes a low-dimensional representation of the input. The model is trained to minimize the reconstruction error, which is the difference between the reconstructed output and ground-truth input.

\begin{figure}
\begin{center}
\includegraphics[width=5cm]{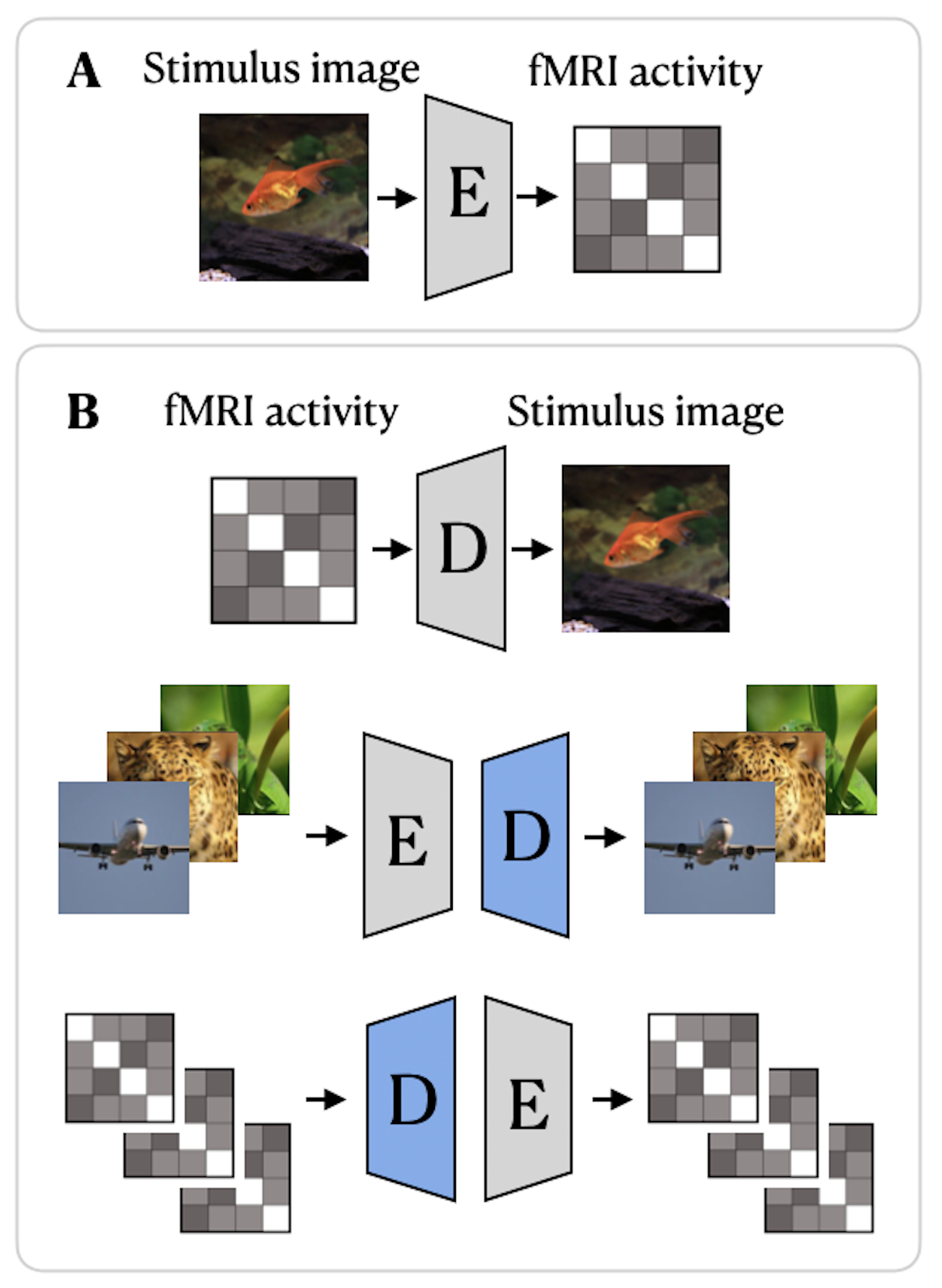}
\end{center}
\caption{\texttt{BeliyEncDec} framework proposed by \cite{beliy_voxels_2019}: \textbf{(A)} supervised training of the Encoder;
\textbf{(B)} supervised and self-supervised training of the Decoder. The weights of the Encoder are fixed. The blue color denotes the components of the model trained on external unlabeled data. The image is adapted from \cite{beliy_voxels_2019}.}
\label{fig:beliy2019}
\end{figure}

%\textbf{\texttt{BeliyEncDec}}. 
\cite{beliy_voxels_2019} presented a CNN-based encoder--decoder model, where the encoder ${E}$ learns the mapping from stimulus images to the corresponding fMRI activity, and a decoder ${{D}}$ learns the mapping from fMRI activity to their corresponding images. The framework of this method, which we refer to as \texttt{BeliyEncDec}, is presented in Figure \ref{fig:beliy2019}. By stacking the encoder and decoder back-to-back, the authors introduced two combined networks ${E}$-${{D}}$ and ${{D}}$-${E}$, whose inputs and outputs are natural images and fMRI recordings, respectively. This allowed the training to be self-supervised on a larger dataset of unlabeled data. Specifically, 50,000 additional images from the ImageNet validation set and test fMRI recordings without stimulus pairs were used as unlabeled natural images and unlabeled fMRI samples. 
The authors demonstrated the advantage of their method by achieving competitive results on two natural image reconstruction datasets: \texttt{Generic Object Decoding} \citep{horikawa_generic_2017} and \texttt{vim-1} \citep{kay_identifying_2008}.
The training was conducted in two steps. In the first step, the encoder ${E}$ builds a mapping from stimulus images to fMRI activity. %acts as a feature extractor. 
It utilizes the weights of the first convolutional layer of the pretrained AlexNet \citep{krizhevsky_imagenet_2012} and is trained in a supervised manner to predict fMRI activity for input images. In the second step, the trained encoder ${E}$ is fixed, and the decoder ${{D}}$ is jointly trained using labeled and unlabeled data. The entire loss of the model consists of the fMRI loss of the encoder ${E}$ and the Image loss (RGB and features loss) of the decoder ${D}$. %and   is computed by optimizing the mean-square error between the stimuli features and reconstruction features extracted from the first and second convolutional layers of VGG-19.

%\textbf{\texttt{GazivEncDec}}. 
In a follow-up study, \cite{gaziv_self-supervised_2020} improved the reconstruction accuracy of \texttt{BeliyEncDec} by introducing a loss function based on the perceptual similarity measure \citep{zhang_unreasonable_2018}. To calculate perceptual similarity loss, the authors first extracted multilayer features from original and reconstructed images using VGG %pretrained on the classification task 
and then compared the extracted features layerwise. To distinguish it from \texttt{BeliyEncDec}, we refer to the framework proposed by \cite{gaziv_self-supervised_2020} as \texttt{GazivEncDec}.

\subsection{Generative methods}
Generative models assume that the data is generated from some probability distribution $p(\mathbf{x})$ and can be classified as implicit and explicit. Implicit models do not define the distribution of the data but instead specify a random sampling process with which to draw samples from $p(\mathbf{x})$. Explicit models, on the other hand,  explicitly define the probability density function, which is used to train the model. 

\textbf{Generative Adversarial Network (GAN).} 

A class of implicitly defined generative models called Generative adversarial networks (GANs) received much attention due to their ability to produce realistic images \citep{goodfellow_generative_2014}. In natural image reconstruction, GANs are widely used to learn the distribution of stimulus images. A GAN %, shown in Figure \ref{fig:dnn_frameworks} B, 
contains generator and discriminator networks. In the image generation task, the generator ${G}$ takes a random noise vector $\mathbf{z}$ (generally sampled from a Gaussian distribution) and generates a fake sample $G(\mathbf{z})$ with the same statistics as the training set images. %The task of the discriminator network ${D}$ is to distinguish the generated fake sample $G(z)$ from the real sample $x$ by maximizing the probability $D(x)$ and minimizing $D(G(z))$. This learning process is formulated as a zero-sum game with the following mini--max loss \citep{goodfellow_generative_2014}:
%\begin{equation}
%    $$E_{x}[\log (D(x))]+E_{z}[\log (1-D(G(z)))]$$
%\end{equation}
%where $E_{x}$ and $E_{z}$ are the expected values of overall real and generated fake samples, G(z). 
During training, the generator's ability to generate realistic images continually improves until the discriminator is unable to distinguish the difference between a real sample and a generated fake one. 
GAN-based frameworks have several desirable properties compared to other generative methods. First, GANs do not require strong assumptions regarding the form of the output probability distribution. Second, adversarial training, which uses the discriminator, allows unsupervised training of the GAN \citep{st-yves_generative_2018}. An illustration of GAN and details on GAN's loss function are provided in Supplementary Material.

%\textbf{\texttt{ShenDNN+DGN}} 
To ensure that reconstructions resemble natural images \cite{shen_deep_2019} further modified their \texttt{ShenDNN} method by introducing a deep generator network (DGN) \citep{dosovitskiy_inverting_2016}. The framework is shown in Figure \ref{fig:shen2019a} B. A DGN, pretrained on natural images using the GAN training process, is integrated with the DNN to produce realistic images, and the optimization is performed on the input space of the DGN. Thus, the reconstructed images are constrained to be in the subspace of the images generated by the DGN. We refer to these framework variations without and with DGN as \texttt{ShenDNN} and \texttt{ShenDNN+DGN} in future references. %Pretrained DGN is available from \url{https://github.com/dosovits/caffe-fr-chairs}. 

\begin{figure}
\begin{center}
\includegraphics[width=\linewidth]{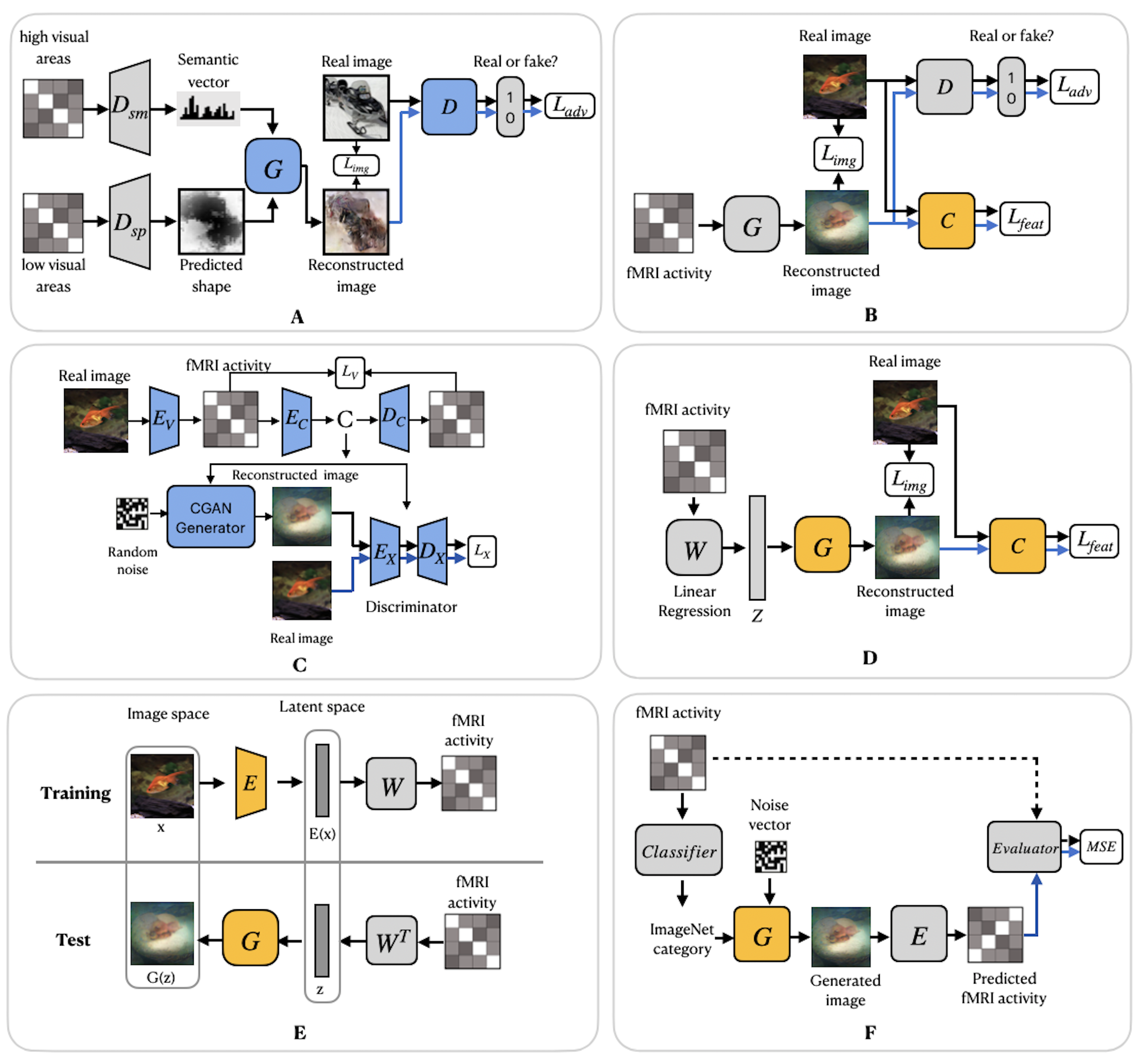}
\end{center}
\caption{GAN-based frameworks.  \textbf{(A)} \texttt{FangSSGAN} framework utilised a semantic decoder $D_{sm}$ and a shape decoder $D_{sp}$. \textbf{(B)} \texttt{ShenGAN} framework introduced a comparator network $C$.  \textbf{(C)} \texttt{StYvesEBGAN} framework consists of three components trained independently: an encoding model $E_V$, denoising autoencoder and $E_C$--$D_C$ and a conditional GAN. \textbf{(D)} \texttt{SeeligerDCGAN} framework based on deep convolutional GAN. \textbf{(E)} \texttt{MozafariBigBiGAN} framework proposed \cite{mozafari_reconstructing_2020}. \textbf{(F)} \texttt{QiaoGAN-BVRM} framework consists of four parts: a classifier, pretrained conditional generator, encoder, and evaluator network. For  \textbf{(A)--(F)}, the pretrained components of the framework are highlighted in yellow. The blue color of the components indicates that they were trained using additional data.}
\label{fig:gan_methods}
\end{figure}

%\textbf{\texttt{FangSSGAN}}. 
Similar to \cite{shen_deep_2019}, \cite{fang_reconstructing_2020} based their work on the finding that visual features are hierarchically represented in the visual cortex.  In the feature extraction step, the authors proposed two decoders, which extract shape and semantic representations from the lower and higher areas of visual cortex. The shape decoder ${D}_{sp}$ is a linear model, and the semantic decoder ${D}_{sm}$ has a DNN-based architecture (Figure \ref{fig:gan_methods} A). In the image reconstruction step, the generator network ${{G}}$ was trained with GAN using the extracted shape and semantic features as conditions for generating the images. We refer to this model as \texttt{FangSSGAN}, where \texttt{SSGAN} stands for the shape and semantic GAN.  The generator ${{G}}$ is a CNN-based network with an encoder--decoder structure \citep{ronneberger_u-net_2015}. To enhance reconstruction quality,  approximately 1,200 additional images, different from those in the training/test set, were sampled from the ImageNet dataset to generate augmented data.
These new images were used to generate shapes and category-average semantic features that were further passed into the GAN. 

%\textbf{\texttt{ShenGAN}}. 
Another GAN-based model was proposed by \cite{shen_end--end_2019}. The end-to-end model directly learns the mapping from fMRI signals to reconstructed images without intermediate transformation or feature extraction (see Figure \ref{fig:gan_methods} B).
%\textcolor{red}{Provided that enough data samples are available, end-to-end training may avoid the potential information loss during the feature extraction step \citep{shen_end--end_2019}}. 
The framework, which we refer to as \texttt{ShenGAN}, was trained using three convolutional  neural networks: 
a generator $G$, a comparator $C$,
and a discriminator $D$. The generator $G$ maps the fMRI data vector $\mathbf{v}$ to $G(\mathbf{v})$, and a discriminator $D$ distinguishes between reconstruction $G(\mathbf{v})$ and the original image $\mathbf{x}$. A comparator network $C$ is pretrained on ImageNet (on image classification task) and used to compare the reconstruction $G(\mathbf{v})$ with the original image  $\mathbf{x}$ by calculating the perceptual loss (similarity in feature space).
The combined loss function is a weighted sum of three terms: loss in image space, perceptual loss and adversarial loss.

The GAN-based methods described so far enhanced the quality of reconstruction by generating more natural-looking images. However, although GANs can generate new plausible samples matching the distribution of samples in the training dataset, they do not allow to control any characteristics of the generated data \citep{jakub_langr_gans_2019}. To solve this issue, \cite{st-yves_generative_2018} implemented the conditional generation of images using a variation of GAN called the energy-based condition GAN or EBGAN \citep{zhao_energy-based_2017}. In their framework, which we refer to as  \texttt{StYvesEBGAN}, the authors first implement the encoding model ${E}_V$ to learn the mapping %$\hat{V} = \boldsymbol{E}_{V}(X)$ 
from stimulus to fMRI, as shown in Figure \ref{fig:gan_methods} C. In addition, \texttt{StYvesEBGAN} utilizes a denoising autoencoder to compress noisy  high-dimensional  fMRI representations into lower-dimensional representations. %$C = \boldsymbol{E}_C(\hat{V})$ 
These  lower-dimensional fMRI representations are further used as a condition vector for the GAN to reconstruct the stimuli. %$Y=\boldsymbol{G}(C)$. 
EBGAN is a more stable framework in terms of training than regular GANs. Instead of a binary classifier, it uses a deep autoencoder network as a discriminator.  %The discriminator acts as an energy function, with energy being the reconstruction error \citep{zhao_energy-based_2017}. 
The authors observed that the reconstruction quality is highly dependent on the voxel denoising autoencoder, which produces a conditioning vector that results in the best reconstruction accuracy.
%The denoiser and generator were pretrained on 32 $\times$ 32 color images from the CIFAR-10 \citep{krizhevsky_learning_2009} dataset.

A group of studies by \cite{seeliger_generative_2018}, \cite{mozafari_reconstructing_2020}, and \cite{qiao_biggan-based_2020} utilized GAN architecture with the assumption that there is a linear relationship  %$Z_{V} = f(V)$
between brain activity and the latent features of GAN. Similar to \texttt{ShenDNN+DGN}, these methods adopted the generator of a pretrained GAN as a natural image prior, which ensures that the reconstructed images follow similar distributions as natural images.

\cite{seeliger_generative_2018} used a deep convolutional GAN (DCGAN) architecture \citep{radford_unsupervised_2016}, which introduced improvements by stacking  convolutional and deconvolutional layers. The authors learn the direct linear mapping from the fMRI space to the latent space  %$Z$ 
of GAN (see Figure \ref{fig:gan_methods} D). For the natural stimuli image domain, the generator ${{G}}$ was pretrained on down-sampled 64 $\times$ 64 converted-to-grayscale images from ImageNet \citep{chrabaszcz_downsampled_2017} and Microsoft COCO \citep{lin_microsoft_2014} datasets. For the handwritten character stimulus domain, DCGAN was pretrained on 15,000 handwritten characters from \citep{maaten_new_2009} and \citep{schomaker_forensic_2000}. Also, a pretrained comparator network ${{C}}$, based on AlexNet, was introduced as a feature-matching network to compute the feature loss $L_{feat}$ across different layers. Overall, the loss is computed as a weighted sum of the pixelwise image loss $L_{img}$ (MAE) and feature loss $L_{feat}$. We refer to this framework as \texttt{SeeligerDCGAN}.

\cite{mozafari_reconstructing_2020} used a variation of GAN, called the BigBiGAN model \citep{donahue_large_2019}, which allowed the reconstruction of even more realistic images. The model generates high-level semantic information due to the BigBiGAN's latent space, which extracts high-level image details from fMRI data.
We refer to this framework as \texttt{MozafariBigBiGAN}. The framework utilizes a pretrained encoder ${E}$ that generates a latent space vector $E(\mathbf{x})$ from the input image $\mathbf{x}$ and generator ${{G}}$ that generates an image $G(\mathbf{z})$ from the latent space vector $\mathbf{z}$ (see Figure \ref{fig:gan_methods} E). During training, the authors computed the linear mapping $W$ from latent vectors $E(\mathbf{x})$ to fMRI activity using a general linear regression model. During the test stage, the linear mapping is inverted to compute the latent vectors $\mathbf{z}$ from the test fMRI activity.%The BigBiGAN model pretrained on ImageNet is available at \url{https://tfhub.dev/deepmind/bigbigan-revnet50x4}. 

%\textbf{\texttt{QiaoGAN-BVRM}}. 
%Despite the improved results, the application of GANs for natural image reconstruction from brain activity might not be satisfactory owing to the small size of the training dataset. 
The GAN-based Bayesian visual reconstruction model (GAN-BVRM) proposed by \cite{qiao_biggan-based_2020} aims to improve the quality of reconstructions from a limited dataset combination and, as the name suggests, uses the combination of GAN and Bayesian learning. From Bayesian perspective, a conditional distribution $p(\mathbf{v}|\mathbf{x})$ corresponds to an encoder which predicts fMRI activity $\mathbf{v}$ from the stimuli image $\mathbf{x}$. On the other hand, an inverse conditional distribution $p(\mathbf{x}|\mathbf{v})$ corresponds to a decoder that reconstructs the stimuli from the fMRI activity. The goal
of image reconstruction is to find the image that has the highest posterior probability $p(\mathbf{x}|\mathbf{v})$,
given the fMRI activity. However, since the posterior distribution is hard to compute, Bayesian theorem is used to combine encoding model $p(\mathbf{v}|\mathbf{x})$ and image prior $p(\mathbf{x})$ through  $p(\mathbf{x} | \mathbf{v}) \propto p(\mathbf{x}) p(\mathbf{v}|\mathbf{x})$. The prior distribution $p(\mathbf{x})$ reflects the predefined knowledge about natural images and is independent of the fMRI activity. 
%using decoded object categories and a pretrained generator. 
The \texttt{QiaoGAN-BVRM} framework is shown in Figure \ref{fig:gan_methods} F, and it consists of four parts: a classifier network, pretrained conditional generator ${{G}}$, encoder ${E}$, and evaluator network. First, a classifier decodes object categories from fMRI data, and then a conditional generator ${{G}}$ of the BigGAN uses the decoded categories to generate natural images. The advantage of the pretrained generator is that it has already learned the data
distribution from more than one million ImageNet natural
images. Therefore, instead of searching the images one by one in a fixed image dataset \citep{naselaris_bayesian_2009}, the generator can produce the optimal image reconstructions that best match with the
fMRI activity via backpropagation. The generated images are passed to the encoder ${E}$, which predicts the corresponding fMRI activity. The proposed 
visual encoding model and the pre-trained generator
of BigGAN do not interfere with each other, which helps to improve the fidelity and naturalness of
reconstruction. The reconstruction accuracy is measured using an evaluator network, which computes the negative average mean squared error (MSE) between the predicted and actual fMRI activity. The reconstructions are obtained by iteratively updating the input noise vector to maximize the evaluator's score. %The authors utilized the generator of BigGAN, pretrained on 128 $\times$ 128 images from the ImageNet dataset, and available from \url{https://github.com/ajbrock/BigGAN-PyTorch}.

\textbf{VAE-GAN. }
The variational autoencoder (VAE) proposed by \cite{kingma_auto-encoding_2014} is an example of an explicit generative network and is a popular generative algorithm used in neural decoding. Similar to autoencoders, the VAE is composed of an encoder and a decoder. But rather than encoding a latent vector, VAE encodes a distribution over the latent space, making the generative process possible.  Thus, the goal of VAE is to find a distribution of the latent variable ${\mathbf{z}}$, which we can sample from $\mathbf{z} \sim q_{\phi}(\mathbf{z}|\mathbf{x})$ to generate new image reconstructions $\mathbf{x}' \sim  p_{\theta}(\mathbf{x}|\mathbf{z})$.  $q_{\phi}(\mathbf{z}|\mathbf{x})$ represents a probabilistic encoder, parameterized with $\phi$, which embeds the input $\mathbf{x}$ into a latent representation $\mathbf{z}$. $p_{\theta}(\mathbf{x}|\mathbf{z})$ represents a probabilistic decoder, parameterized with $\theta$, which produces a distribution over the corresponding $\mathbf{x}$. The details on VAE and its loss function are provided in Supplementary Material.

%The critical feature that distinguishes VAE from other autoencoders is that the latent spaces are continuous. This allows easy random sampling and interpolation, that is, replicating the output and generating images similar to the input. 
A hybrid model by \cite{larsen_autoencoding_2016} integrates both the VAE and GAN in a framework called VAE-GAN. VAE-GAN combines VAE to produce latent features and GAN discriminator, which learns to discriminate between fake and real images. In  VAE-GAN, the VAE decoder and GAN generator are combined into one. The advantages of VAE-GAN are as follows. First, the  GAN's adversarial loss enables generating visually more realistic images. Second, VAE-GAN achieves improved stability due to VAE-based optimization. This helps to avoid mode collapse inherent to GANs, which refers to a generator producing a limited subset of different outcomes \citep{ren_reconstructing_2021, xu_learning_2021}. 

%Figure \ref{fig:dnn_frameworks} D illustrates the VAE-GAN framework.
%The VAE-GAN loss is computed as follows:
%\begin{equation}
%\label{eq:vae-gan}
%    $$L_{VAE-GAN} = L_{Dis} + L_{KL}(q_{\theta}(z|x)|| p(z)) + L_{GAN}$$.
%\end{equation}
%Compared to the VAE loss in equation \ref{eq:vae}, VAE-GAN loss includes $L_{GAN}$ loss, and the reconstruction loss $L(x, y)$ is replaced with a reconstruction error expressed using the intermediate features of the GAN discriminator $L_{Dis}$.

\begin{figure}[h]
\begin{center}
\includegraphics[width=9cm]{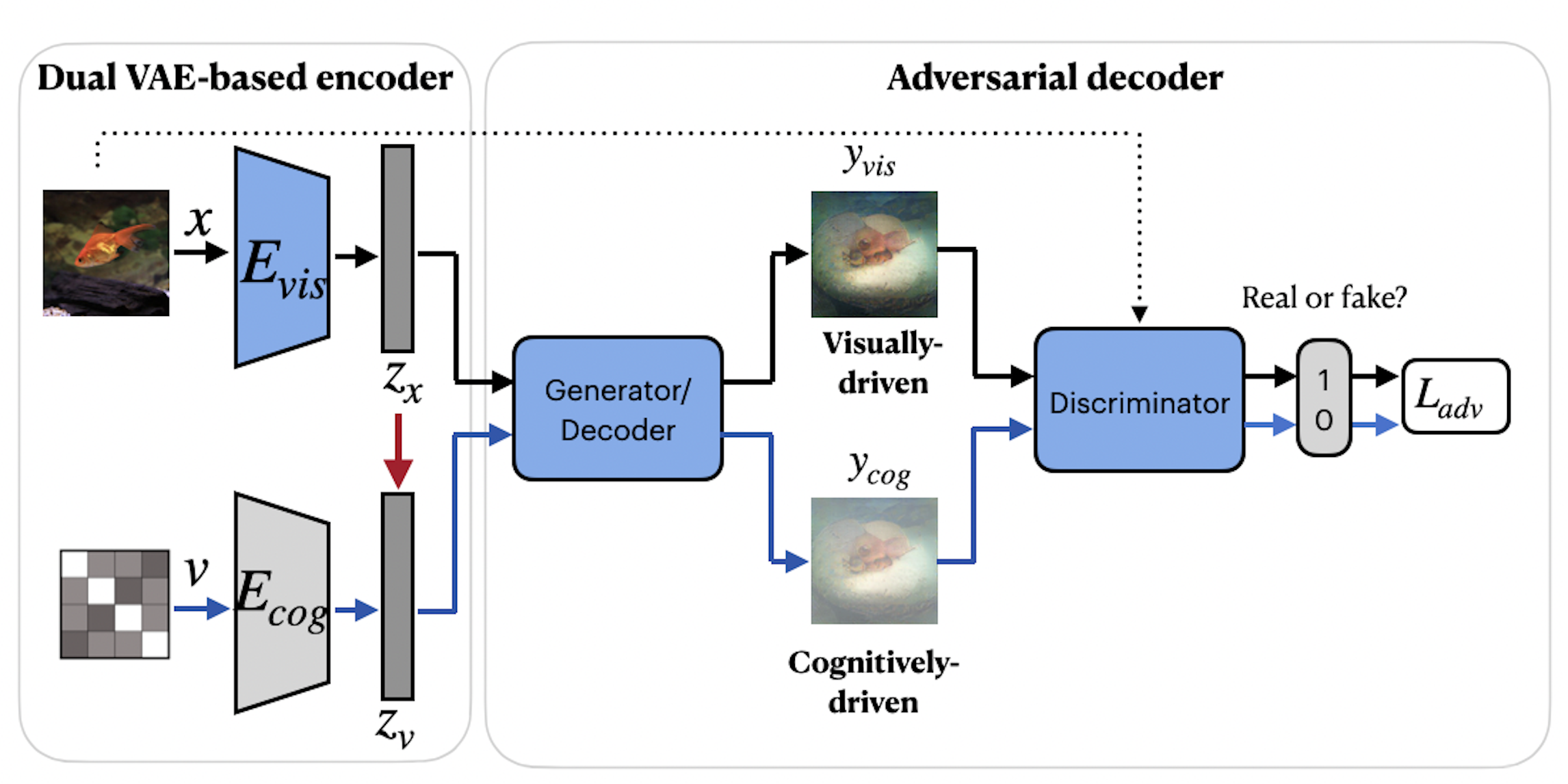}
\end{center}
\caption{\texttt{RenD-VAE/GAN} framework consists of three main components: dual VAE-based encoder, adversarial decoder, and discriminator. The visual encoder $E_{vis}$, cognitive encoder $E_{cog}$, generator, and discriminator were used during the training. During testing, only a trained cognitive encoder %visual decoder 
and generator were used for the inference. The red arrow denotes the transfer of knowledge from the teacher network $E_{vis}$ to the student network $E_{cog}$. The components in blue denote training on external unlabeled natural images (without fMRI activity) from ImageNet, which do not overlap with images in the train/test set.
}
\label{fig:ren2021}
\end{figure}

A group of studies on reconstructing natural images from brain activity patterns, including %Gucluturk et al. \citep{gucluturk_reconstructing_2017}, %Du et al.
%\citep{du_reconstructing_2019}, 
\cite{ren_reconstructing_2021} and \cite{vanrullen_reconstructing_2019}, incorporated probabilistic inference using VAE-GAN. % \citep{vanrullen_reconstructing_2019}.\citep{gucluturk_reconstructing_2017}. 
%\textcolor{red}{\textbf{GucluturkNNN.}}Gucluturk et al. \citep{gucluturk_reconstructing_2017} combined probabilistic inference with a GAN training process to reconstruct face images. For the encoder model, the authors used pretrained VGG-16 \citep{simonyan_very_2015} and VGG-Face \citep{parkhi_deep_2015} pretrained for face recognition. \textcolor{red}{add more info}.
%\textbf{\texttt{RenD-VAE/GAN}}. 
In a recent work by \cite{ren_reconstructing_2021}, the authors presented a combined network called 
Dual-Variational Autoencoder/ Generative Adversarial Network (D-VAE/GAN). The framework, which we named \texttt{RenD-VAE/GAN}, consists of a dual VAE-based encoder and an adversarial decoder, as illustrated in Figure \ref{fig:ren2021}. Dual-VAE consists of two probabilistic encoders: visual $E_{vis}$ and cognitive $E_{cog}$, which encode stimuli images $\mathbf{x}$ and brain activity patterns $\mathbf{v}$ to corresponding latent representations $\mathbf{z_x}$ and $\mathbf{z_v}$. 
The framework is trained in three sequential stages.
In the first stage, visual stimuli images %$X$ 
are used to train the visual encoder $E_{vis}$, generator ${{G}}$, and discriminator ${{D}}$. $E_{vis}$ learns the direct mapping from visual images into latent representations. %$Z_{X}=E_{vis}(X)$. 
Then, using output of $E_{vis}$, % $Z_{X}$, 
the generator ${{G}}$ is trained to predict the images %${Y_X} = G(Z_{X})$, 
and ${{D}}$ is trained to discriminate the predicted images %${Y_X}$ 
from real images. %$X$. 
In the second stage, $E_{cog}$ is trained to map high-dimensional fMRI signals %$V$ 
to cognitive latent features. % $Z_{V} = E_{cog}(V)$. 
The generator ${{G}}$ is fixed and ${{D}}$ is trained to discriminate between the stimuli images %$Y_X$ 
produced in the first stage and the cognitively-driven reconstructions from cognitive latent features. % $Y_V=G(E_{cog}(V))$. 
This way, $E_{cog}$ is forced to generate visual and cognitive latent representations similar to each other. %$Z_V$similar to $Z_X$. 
In the last training stage, $E_{cog}$ is fixed, whereas ${{G}}$ and ${{D}}$ are fine-tuned on fMRI signals %$V$ 
to improve the accuracy of the generated images via cognitive latent representations. %$Y=G(E_{cog}(V))$. 
In this stage, ${{D}}$ is trained to discriminate between real stimuli images and reconstructed images. During testing, only a trained cognitive encoder %visual decoder 
and generator were used for the inference.
Since $E_{vis}$ takes visual stimuli as input, its learned latent representations $\mathbf{z_x}$ can guide $E_{cog}$ to learn the latent representations $\mathbf{z_v}$. Thus, in the second training stage, the authors implement the concept of knowledge distillation by transferring knowledge from $E_{vis}$ to $E_{cog}$, which together represent the teacher and student networks \citep{hinton_distilling_2015}. The learned latent representation vectors significantly improve the reconstruction quality by capturing visual information, such as color, texture, object position, and attributes. %The framework, which we named \texttt{RenD-VAE/GAN}, was pretrained on 10,000 randomly sampled unlabeled images from ImageNet (images without fMRI recordings) which did not overlap with the images presented in the \texttt{Generic Object Decoding} dataset.

\begin{figure}[h]
\begin{center}
\includegraphics[width=8cm]{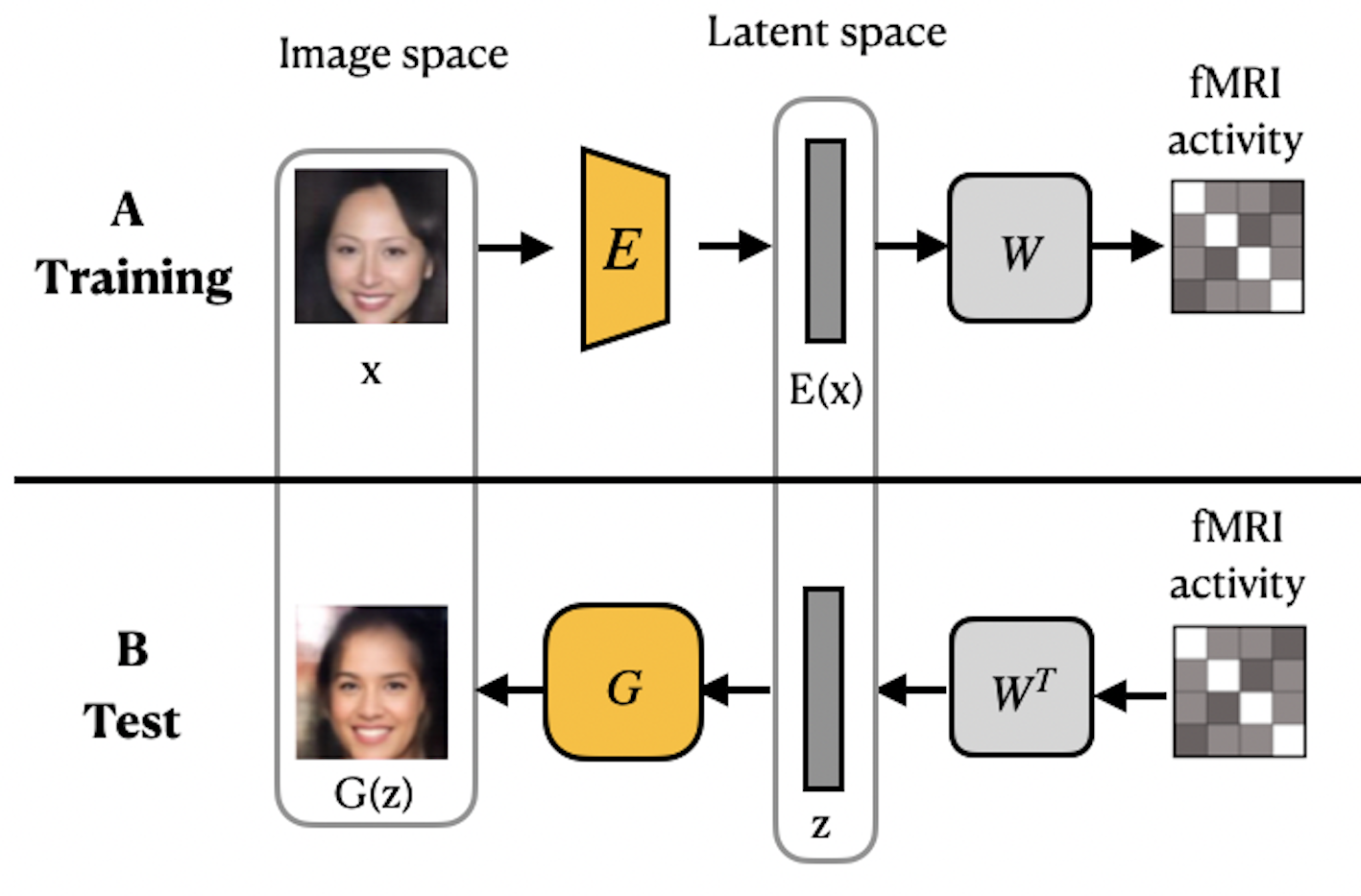}
\end{center}
\caption{\texttt{VanRullenVAE-GAN} framework proposed by \cite{vanrullen_reconstructing_2019}. The encoder $\boldsymbol{E}$ maps a stimulus image onto the latent representation $\mathbf{z}$. The generator ${G}$ uses $\mathbf{z}$ to reconstruct the stimuli image. The pretrained components are shown in yellow.}
\label{fig:vanRullen2019}
\end{figure}

%\textbf{\texttt{VanRullenVAE–GAN}}. 
\cite{vanrullen_reconstructing_2019} utilized VAE network pretrained on CelebA dataset using GAN procedure to learn variational latent space. Similar to \texttt{MozafariBigBiGAN} framework, the authors learned a linear mapping between latent feature space and fMRI patterns, rather than using probabilistic inference \citep{gucluturk_reconstructing_2017}. In the training stage, the pretrained encoder from VAE-GAN is fixed and  the linear mapping between latent feature space and fMRI patterns is learned. For the test stage, fMRI patterns are first translated into VAE latent codes via inverse mapping, and then these codes are used to reconstruct the faces. The latent space of a VAE is a variational layer that provides a meaningful
description of each image and can represent faces and facial features as linear combinations of each other. Owing to the training objective of the VAE,  the points which appear close in this space are mapped onto similar face images, which are always visually
plausible. Therefore, the VAE's latent space ensures that the brain
decoding becomes more robust mapping errors.  As a result, the produced reconstructions from VAE-GAN appear to be more realistic and closer to the original stimuli images. This method not only allows the reconstruction of naturally looking faces but also decodes face gender.
%The VAE–GAN model is pretrained on a large dataset of celebrity faces from the CelebA dataset \citep{liu_deep_2015}. 
In terms of architecture, the framework, which we refer to as \texttt{VanRullenVAE-GAN},  consists of three networks, as shown in Figure \ref{fig:vanRullen2019}.

\section{Reconstruction evaluation} \label{sec:mertics}
\begin{figure*}
\begin{center}
\includegraphics[width=14cm]{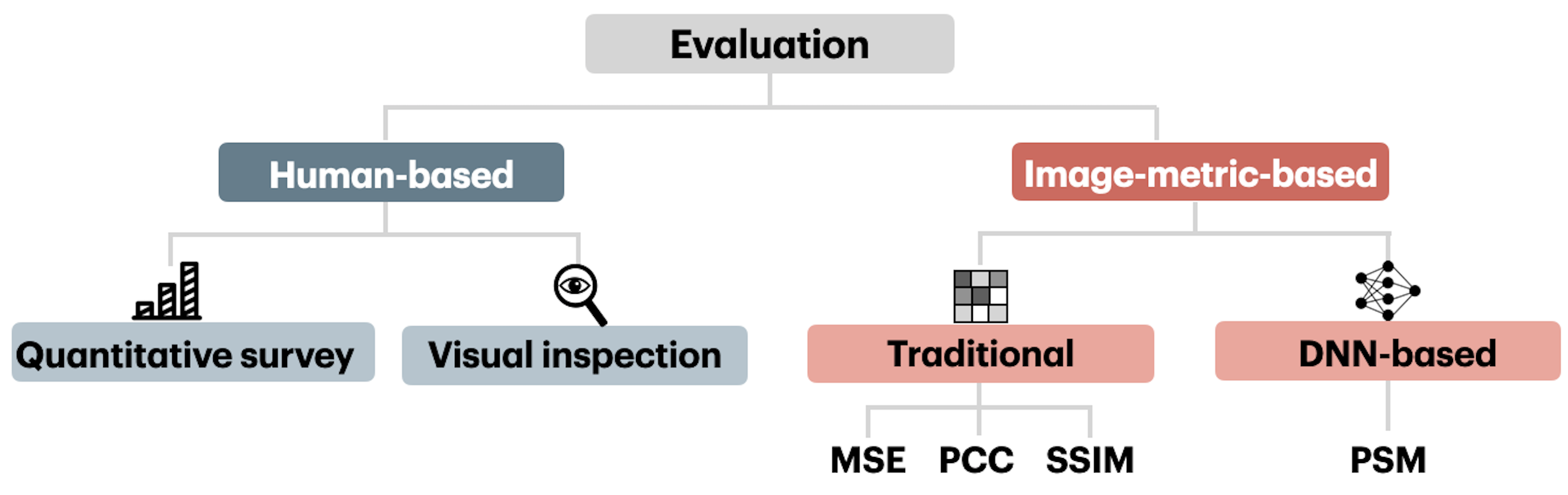}
\end{center}
\caption{Image-metric-based and human-based evaluation.}
\label{fig:eval_metrics}
\end{figure*}

The evaluation of reconstruction methods is based on human-based and image metrics, which we schematically present in Figure \ref{fig:eval_metrics}. %provides an overview of the image-metric-based and human-based evaluation reviewed in this study. %An additional challenge is that no unified set of metrics exist for evaluation. From Table \ref{table:metrics}, it can be seen that the existing methods substantially vary in terms of the used image comparison settings and the underlying evaluation metrics. 
We first present human-based and image metrics and then describe the differences in image comparison settings. %Each evaluation approach has its strengths and weaknesses. 

\begin{table*}[]
    \centering
    \caption{Comparison of methods in terms of the used evaluation metrics. PCC stands for the Pearson correlation coefficient.}
    \label{table:metrics}
\resizebox{12cm}{!}{    
\begin{tabular}{lcccc}
\hline
 &
  \multicolumn{2}{c}{\textbf{Human-based metrics}} &
  \multicolumn{2}{c}{\textbf{Image metrics}} \\ \hline
\multicolumn{1}{c}{\textbf{Reference}} &
  \textbf{\begin{tabular}[c]{@{}c@{}}Quantitative \\ survey\end{tabular}} &
  \textbf{\begin{tabular}[c]{@{}c@{}}Visual \\ inspection\end{tabular}} &
  \textbf{Traditional} &
  \textbf{PSM} \\ \hline
\cite{seeliger_generative_2018} &
  \cmark &
  \cmark &
  \xmark &
  \xmark \\ \hline
\cite{shen_deep_2019} &
  \cmark &
  \cmark &
  Pairwise PCC &
  \xmark \\ \hline
\cite{shen_end--end_2019} &
  \cmark &
  \cmark &
  \begin{tabular}[c]{@{}c@{}}Pairwise PCC\\ Pairwise SSIM\end{tabular} &
  \xmark \\ \hline
\cite{beliy_voxels_2019} &
  \xmark &
  \cmark &
  2,5,10-way PCC &
  \xmark \\ \hline
\cite{gaziv_self-supervised_2020} &
  \cmark &
  \cmark &
  \xmark &
  2,5,10-way PSM \\ \hline
\cite{qiao_biggan-based_2020} &
  \xmark &
  \cmark &
  \begin{tabular}[c]{@{}c@{}}PCC\\ SSIM\end{tabular} &
  AlexNet \\ \hline
\cite{fang_reconstructing_2020} &
  \xmark &
  \cmark &
  Pairwise PCC &
  \xmark \\ \hline
\cite{mozafari_reconstructing_2020} &
  \xmark &
  \cmark &
  \begin{tabular}[c]{@{}c@{}}Pairwise PCC\\ Pix-Comp (2-way PCC)\end{tabular} &
  Inception-V3 \\ \hline
\cite{ren_reconstructing_2021} &
  \cmark &
  \cmark &
  \begin{tabular}[c]{@{}c@{}}Linear correlation\\ SSIM\\ 2,5,10-way PCC\end{tabular} &
  \xmark \\ \hline
\end{tabular}
}
\end{table*}

\subsection{Human-based evaluation}
The intuitive method of measuring the quality of reconstruction in natural image reconstruction task is by employing human evaluators.
Human-based evaluation can be conducted quantitatively
and qualitatively through visual inspection.
%The most straightforward method of measuring the quality of reconstruction is by employing human assessment, where human subjects judge how good are the reconstructed images.  

%\subsubsection{Quantitative survey}
For quantitative human-based assessment, a behavioral study involving human subjects is conducted. In this study the reconstructed image is compared to the original or several candidate images, containing the original image.  % and two candidate images 
From the given candidate images, subjects are instructed to choose the one that appears to have a higher resemblance to the original. Such behavioral studies can be conducted by employing human evaluators or using Amazon Mechanical Turk\footnote{\url{www.mturk.com}} \citep{seeliger_generative_2018, gaziv_self-supervised_2020}. 

%\subsubsection{Qualitative visual inspection/comparison}
Owing to the additional time and human input required for human-based evaluation, several recent studies omit quantitative human evaluation in favor of qualitative visual inspections. For visual comparison, the set of original images and their reconstructions from different reconstruction methods are juxtaposed for ease of comparison (see Figures \ref{fig:vis_comp} A and  \ref{fig:vis_comp} B). Reconstructions are usually compared in terms of image sharpness/blurriness, matching shapes, colors, and low/high-level details. Many recent works focus on emphasizing the ``naturalness'' of their reconstructions, despite the reconstructions deviating significantly from the actual images in terms of the object category (see reconstructions in column 4 in Figure \ref{fig:vis_comp} B for example).

Although human-based evaluation is a more reliable measurement of the quality of the reconstructed image, it suffers from the following limitations. First, human-based evaluation is time consuming and expensive because the process requires a well-designed evaluation study and the recruitment of human subjects.
Second, the results can be heavily affected by subjects' physical and emotional conditions or external conditions, such as lighting or image display \citep{wang_image_2004, rolls_invariant_2012}. Table \ref{table:metrics} shows that only several studies conducted the quantitative human-based evaluation.

%\textcolor{red}{However, human-based evaluation is not practical in real-world applications because of the high cost and additional time requirements. In addition, human-based evaluation is subjective, as the results can be obscured by external factors, such as the subjects' vision, mood, prior information of the image content, viewing angle, and distance. %Therefore, it is more practical to use objective methods that can fairly assess the quality of image reconstruction.

\subsection{Image-metric-based evaluation}%{Objective evaluation}

As an alternative to human-based evaluation, image-metric-based evaluation is used to accurately and automatically assess image reconstruction quality. The use of image metrics for evaluation is more practical, and unlike human-based assessment, is unbiased towards external factors. However, the image-metric-based evaluation can provide only an approximation of the visual comparison mechanism inherent to a human subject, and thus are far from being perfect \citep{wang_image_2004}. %The reviewed metrics include 1) traditional metrics, which allow low-level pixelwise comparison, and 2) higher-level DNN-based metric, which compares images on multiple levels of abstraction. 

Nowadays, there exist various image metrics that can compare images at different levels of perceptual representation. %That is, the reconstruction quality is be measured either in terms of low-level perceptual similarity on pixel level (pixel colors, edges, and shapes) or using high-level perceptual similarity (object attributes and category). 
Image metrics used in the visual decoding literature can be categorized into traditional metrics that capture low-level perceptual similarities and more recent ones that capture high-level perceptual similarity. The conventional metrics, which include the mean squared error (MSE), pixelwise Pearson correlation coefficient (PCC), structural similarity index (SSIM), and their variants, are computed in pixel space and capture low-level perceptual similarity. The metric that captures high-level perceptual similarity relies on multilevel feature extraction from DNN and can compare images at a higher level of perceptual representation. The high-level metric we considere here is called Perceptual Similarity Metric (PSM).

\textbf{MSE} is the simplest traditional metric for assessing image reconstruction quality. Given $x_i$ and $y_i$, which are the flattened one-dimensional representations of the original and the reconstructed images, %Given a set of $N$ flattened one-dimensional representations of original images $X = \{x_1, x_2, ..., x_N\}$ and their corresponding reconstructed images $Y = \{y_1, y_2, ..., y_N\}$, the 
the MSE estimated over $N$ samples is computed as
\begin{equation}
    MSE = \frac{1}{N}\sum_{i=1}^N(x_i - y_i)^2.
    \label{eq:label}
\end{equation}
Several characteristics of MSE, including simplicity of implementation and fast computation,  make it a widely used performance metric in signal processing. However, MSE shows poor correspondence to human visual perception, due to some of the underlying assumptions: MSE is independent of the spatial relationship between image pixels and considers each of them to be equally important \citep{wang_image_2004}.

\textbf{PCC} is widely used in statistical analysis to measure the linear relationship between two variables. The following equation is used to compute the pixelwise Pearson correlation between the flattened 1-D representations of the original image $x$ and the reconstructed image $y$:
\begin{equation}
    PCC(x, y) = \frac{\sum(x - \mu_{x})(y - \mu_{y})}{\sqrt{\sum(x - \mu_{x})^2 \sum(y - \mu_{y})^2}},
\label{eq:ppc}
\end{equation}
where $\mu_{x}$ and $\mu_{y}$ are the mean intensities of the flattened one-dimensional vectors $x$ and $y$, respectively. PCC is the most common metric used across the surveyed works (see Table \ref{table:metrics}), with slight variations in naming and implementation: Pairwise PCC \citep{shen_deep_2019, shen_end--end_2019}, pixel correlation \citep{ren_reconstructing_2021}, Pix-Comp \citep{mozafari_reconstructing_2020}, and n-way PCC \citep{beliy_voxels_2019}. The limitation of PCC is its sensitivity to changes in the edge intensity or edge misalignment. Thus, the metric tends to assign higher scores to blurry images than to images with distinguishable but misaligned shapes \citep{beliy_voxels_2019}.
 
\textbf{SSIM} is widely used image similarity metric that captures structural information from images.
Wang et al. proposed SSIM as a quality assessment metric that resembles the characteristics of the human visual perception \citep{wang_image_2004}.  Unlike PCC, which treats each pixel of the image independently, SSIM measures the similarity of spatially close pixels between the reconstructed and original images.  Given two images, SSIM is computed as a weighted combination of three comparative measures: luminance, contrast, and structure. 
Assuming an equal contribution of each measure, the SSIM is first computed locally between the corresponding windows $p$ and $q$ of images $x$ and $y$: 
\begin{equation}
    SSIM(p, q) = \frac{(2\mu_p \mu_q + C_1)(2\sigma_{pq} + C_2)}{(\mu_p^2 +  \mu_{q}^2 + C_1)(\sigma_p^2 +  \sigma_{q}^2 + C_2)} ,
    \label{eq:ssim}
\end{equation}
where $\mu_{p}$ and $\mu_{q}$ are the mean intensity values of $p$ and $q$, respectively; $\sigma_{p}^{2}$ and $\sigma_{q}^{2}$ are  the variances of $p$ and $q$, respectively; $\sigma_{pq}$ is the covariance of $p$ and $q$, and $C_{1}$ and $C_{2}$ are constants that ensure stability when the denominator is close to zero.
The global SSIM score is computed as the average of all $M$ local SSIM scores:
\begin{equation}
    SSIM(x, y) = \sum_{i=1}^{M}SSIM(p_i, q_i).
    \label{eq:ssim_global}
\end{equation}
%The complete implementation of the SSIM measure is available as part of an open-source image processing \textit{scikit-image} library for Python.

\textbf{PSM.} \label{sec:psm}
Despite the wide adoption of SSIM as a perceptual metric, it compares poorly with many characteristics of human perception \citep{zhang_unreasonable_2018}. Several studies, including \cite{gucluturk_reconstructing_2017}, \cite{qiao_accurate_2018}, \cite{mozafari_reconstructing_2020}, and \cite{gaziv_self-supervised_2020}, emphasize the importance of higher-level perceptual similarity over lower-level metrics in evaluation because of the better correspondence of higher-level perceptual similarity to human perceptual judgments \citep{zhang_unreasonable_2018}. 

As the general principle, a CNN is used for extracting hierarchical multilayer features of input images, which are further compared across layers using a distance metric of choice. However, the definition and implementation of the perceptual similarity metric in terms of the distance metric or feature extraction network vary across studies. For example, \cite{qiao_biggan-based_2020} utilized five convolutional layers of the AlexNet \citep{krizhevsky_imagenet_2012} to extract hierarchical features. The other study by \cite{mozafari_reconstructing_2020}, proposed a high-level similarity measure, which measures perceptual similarity based on the output of only the last layer of Inception-V3 \citep{szegedy_rethinking_2016}. Finally, \cite{gaziv_self-supervised_2020} used the PSM definition proposed in \citep{zhang_unreasonable_2018} with the pretrained AlexNet with linear calibration. Following \cite{gaziv_self-supervised_2020}, we provide a PSM definition by \cite{zhang_unreasonable_2018} in the following equation:
\begin{equation}
\label{eq:psm}
    d\left(x, y\right)=\sum_{l} \frac{1}{H_{l} W_{l}} \sum_{h, w}\left\|w_{l} \odot\left({f}_{x}^{l}-{f}_{y}^{l}\right)\right\|_{2}^{2},
\end{equation}
where $d\left(x, y\right)$ is the distance between the original image $x$ and the reconstructed image $y$. ${f}_{x}^{l}, {f}_{y}^{l}$ %$\in$ $\mathbb{R}^{H_{l} \times W_{l} \times C_{l}}$
represent layerwise activations normalized across channels for layer $l$. 
The activations are scaled channelwise by vector $w_{l} \in \mathbb{R}^{C_{l}}$, spatially averaged, and summed layerwise. 
%The implementation of PSM by \citep{zhang_unreasonable_2018} is available at \url{https://github.com/richzhang/PerceptualSimilarity}.

Note that the underlying CNN model used for computing the PSM should be selected cautiously. Because many studies use pretrained CNN models, it is important to avoid using the same model for both training and evaluation, which may lead to a potential bias in evaluation. For example, several methods, including \cite{shen_deep_2019} and \cite{beliy_voxels_2019}, used VGG-19 \citep{simonyan_very_2015} for pretraining. Therefore, the VGG-19 model should not be used for evaluation, as the objective of evaluation and optimization functions would be the same, and the evaluation would produce a higher similarity between original and reconstructed images. 

\subsection{Image comparison setting}
%There are multiple ways for comparing the images. 
We describe three image comparison settings existing in literature: %, which determine how the images are sampled for image-metric-based or human-based evaluation: 
1) one-to-one comparison, 2) pairwise comparison, and 3) $n$-way comparison. Each of these comparison settings can work with any image or human-based metric of choice. %and qualitative human-based metric (confirm wording). [Similar to the image-metric-based evaluation, any of the image comparison settings can be applied to measure the quality of reconstructed images \citep{shen_end--end_2019, shen_deep_2019}.]}

\textbf{One-to-one} is the simplest comparison setting which computes the similarity score of a reconstruction against ground truth using the given metric, for example, MSE or PCC. %\textcolor{red}{For example, Shen et al. used MSE? metric, and others used other metric.} 
However, the absolute values of qualitative metrics  computed only on a single pair of original and reconstructed images are challenging to interpret \citep{beliy_voxels_2019}. Therefore, pairwise similarity and $n$-way identification are often used to measure the reconstruction quality across the dataset. 

\textbf{Pairwise comparison} %\label{sec:pairwise}
analysis is performed by comparing a reconstructed image with two candidate images: the ground-truth image and the image selected from the remaining set, resulting in a total of $N(N-1)$ comparisons:
%Given a set of $N$ original images $X = \{x_1, x_2, ..., x_N\}$ and the corresponding reconstructions  $Y = \{y_1, y_2, ..., y_N\}$, the pairwise comparison score  is computed as
\begin{equation}
    score = \frac{1}{N(N-1)}\sum_{i=1}^{N}\sum_{\substack{j=1\\j \neq i}}^{N} \sigma\left(m\left(y_{i}, x_{i}\right), m\left(y_{i}, x_{j}\right)\right),
\label{eq:pairwise_ppc}
\end{equation}
where $m$ is the metric of interest and 
\begin{equation}
    \sigma(a, b)=\left\{\begin{array}{ll}
    1 & a>b \\
    0 & \text { otherwise }
    \end{array}\right.
    \label{eq:sigma}
\end{equation}
The trial is considered correct if the metric score of the reconstructed image with the corresponding stimulus image is higher than that with the nonrelevant stimulus image. For metrics that imply that the lower, the better (such as MSE), the expression in the equation \ref{eq:sigma} is modified to find the smallest value. Finally, the percentage of total correct trials is computed as the ratio of correct trials among all trials \citep{shen_deep_2019, shen_end--end_2019, beliy_voxels_2019}. The chance-level accuracy is $50\%$.

%\subsubsection{$n$-way identification task} \label{sec:n-way}
In \textbf{$\boldsymbol{n}$-way identification} 
%is another method that allows the evaluation of the reconstruction accuracy across the dataset. In $n$-way identification, 
each reconstructed image is compared to $n$ randomly selected candidate images, including the ground truth. Several studies, including \cite{beliy_voxels_2019} and \cite{ren_reconstructing_2021}, used $n=2,5,10$ for the $n$-way identification accuracy computed using PCC. In a more recent work, \cite{gaziv_self-supervised_2020} report $n=2,5,10,50$-way identification accuracy based on PSM. An addition source of confusion is the absence of naming conventions: \cite{ren_reconstructing_2021} and \cite{mozafari_reconstructing_2020} referred to $n$-way identification accuracy computed with PCC as Pixel Correlation and pix-Comp, respectively.

\section{Fair comparison across the methods} \label{sec:comp_eval}
For fair comparison of the existing methods, we chose those that satisfied one of the following criteria: 1) the availability of the complete code for reproducing the results and 2) the availability of reconstructed images for running the evaluation. This allowed us to compare five state-of-the-art methods on the \texttt{DIR} dataset, both visually (Section \ref{sec:vis_comp}) and quantitatively (Section \ref{sec:quant_comp}). For the \texttt{GOD}, because of the lack of a complete set of reconstructions for the chosen methods, we only present a visual comparison in Section \ref{sec:vis_comp}. Visual comparison for \texttt{vim-1} datasets is provided in Supplementary Material.

Our analysis of recent works on natural image reconstruction reveals that only a few comply with good machine learning practices regarding the fairness of evaluation. Unfair evaluation can be reflected in the comparison across different datasets, selecting specific subjects in reporting the results, and discrepancies in using the evaluation metrics. This motivated us to perform a rigorous empirical evaluation of the methods, i.e. % with the following parameters standardized: 1) 
\textit{cross-subject} %and per-subject 
evaluation across \textit{common} metrics using a \textit{common} dataset.

\textbf{Evaluation on a common dataset}.
To standardize the objective evaluation process, we perform the
quantitative assessment on the \texttt{DIR} dataset for methods that we found to be reproducible, that is, \texttt{ShenDNN}, \texttt{ShenDNN+DGN}, \texttt{ShenGAN}, and \texttt{BeliyEncDec}. For \texttt{FangSSGAN}, we ran an evaluation based on the reconstructions provided by the authors. 

It is important to distinguish the five-subject \texttt{GOD} dataset \citep{horikawa_generic_2017} from the three-subject \texttt{DIR} dataset \citep{shen_deep_2019}, which uses the same stimuli images but is quite different in terms of the image presentation experiment and characteristics of fMRI activity data. Our choice of the \texttt{DIR} as a common natural image dataset is due to the following reasons. First, unlike the similar \texttt{GOD} dataset, \texttt{DIR} was acquired specifically for the natural image reconstruction task and contains a larger number of training samples due to increased number of repeats in image presentation experiment. 
In addition, this dataset might be of interest for studying the generalizability of natural image reconstruction methods to artificial shapes, which we describe in detail in Supplementary Material.% \ref{sec:generalization}. 
 
When training \texttt{ShenDNN}, \texttt{ShenDNN+DGN}, and \texttt{BeliyEncDec} on the \texttt{DIR}, we used the original training settings. For \texttt{ShenGAN}, we used the pretrained model provided by the authors. To maximize the amount of training data, each presented stimulus sample was treated as an individual sample  \citep{shen_end--end_2019}. For reconstruction, we averaged the test fMRI activations across trials corresponding to the same stimulus to increase the signal-to-noise ratio. This resulted in 6,000 training and 50 test fMRI-image samples. Note that \texttt{BeliyEncDec} was initially implemented for \texttt{GOD} dataset. For \texttt{BeliyEncDec}, averaging both training and test fMRI samples across the repeats resulted in the best performance. This confirms with the authors' observation that an increased number of fMRI repeats results in improved reconstruction \citep{beliy_voxels_2019}. Additionally, we normalized training fMRI vectors to have a zero mean and unit standard deviation. The mean and standard deviation of the training fMRI data were used to normalize the test fMRI data.

\textbf{Evaluation across common metrics.}
We perform the evaluation on natural images %, shapes, and letters 
from the \texttt{DIR} based on MSE, PCC, SSIM, and PSM metrics described in Section \ref{sec:mertics}.
%An additional challenge is that no unified set of metrics exist for evaluation. From Table \ref{table:metrics}, it can be seen that the existing methods substantially vary in terms of the used image comparison settings and the underlying evaluation metrics. 
We notice that there is no consensus among recent works on a standard set of evaluation metrics (see Table \ref{table:metrics}). Moreover, several studies introduce new evaluation metrics or use variations of existing metrics, potentially more favorable for their results. In contrast, we present an evaluation of the methods across all the image metrics used in the related methods.

It is also important to note that different methods generate output images of various sizes due to memory restrictions and variations in the pretrained model (we refer to Supplementary Material %\ref{table:comp_resolution} 
for details on output image resolutions). 
The evaluation metrics can be sensitive to the size of the image and the choice of upscaling or downscaling algorithms. 
For fairness, we rescaled the reconstructions for the \texttt{DIR} to the common size and use a bicubic algorithm for image resizing. We evaluated the reconstructed images using a resolution of 256 $\times$ 256 pixels, which is the highest among the chosen methods. For methods with a lower reconstruction image size, we applied image upscaling.%ed the images to 256 $\times$ 256 pixels.

\subsection{Visual comparison results} \label{sec:vis_comp}

\begin{figure*}[h]
\begin{center}
    \includegraphics[width=\linewidth]{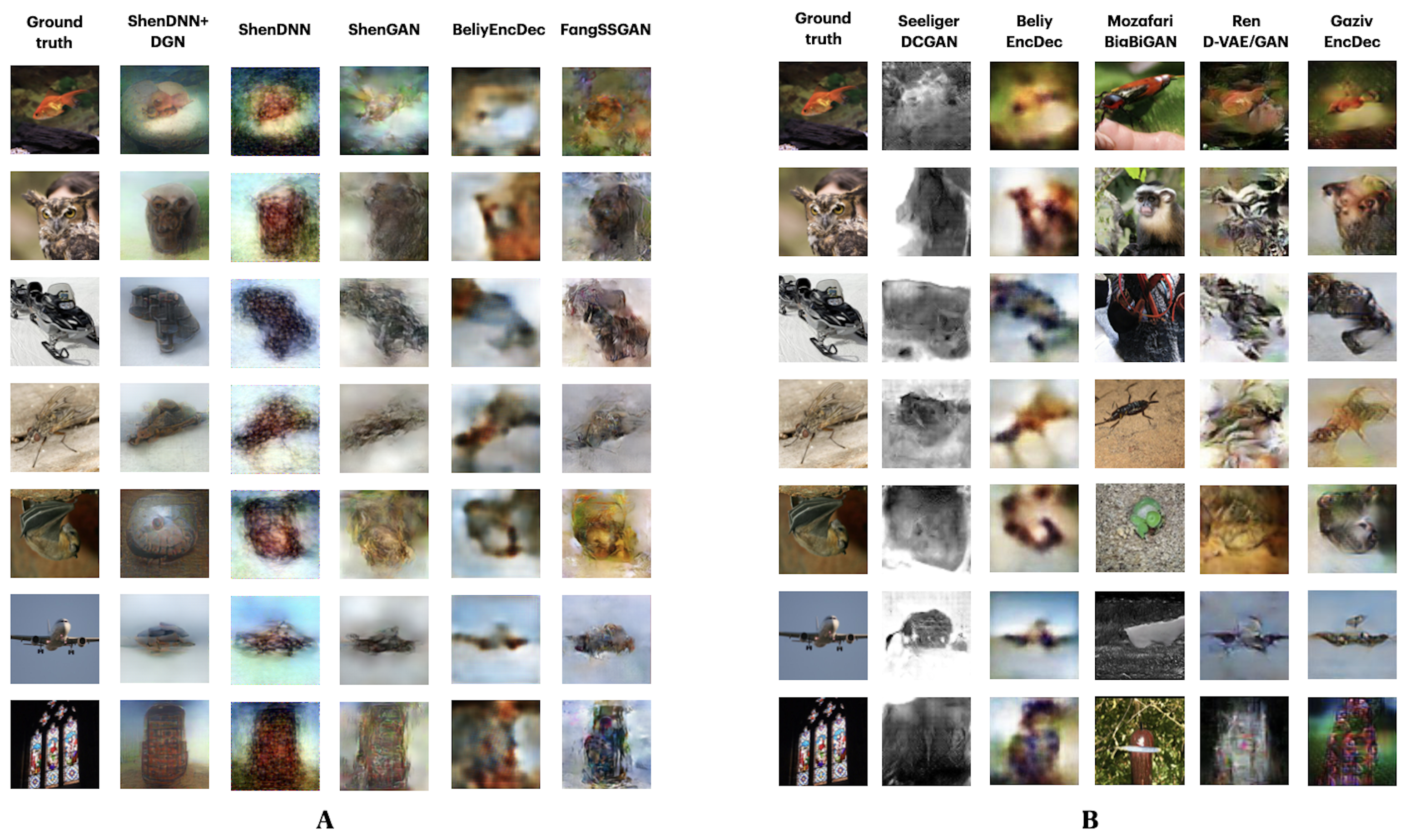}
\end{center}
\caption{
\textbf{(A)} Visual comparison of methods on \texttt{Deep Image Reconstruction} dataset for subject 1. The reconstructions for all methods except for  \cite{fang_reconstructing_2020} are obtained by reproducing the experiments. 
For \textbf{(A)} and \textbf{(B)}, the stimulus images are shown in the first column. The corresponding reconstructed images from each method are shown in the subsequent columns. 
\textbf{(B)} Visual comparison of the methods on the \texttt{GOD} dataset. Due to the unavailability of complete reconstruction data for \texttt{GOD}, visual reconstructions correspond to the same image stimuli but different subjects. For \texttt{BeliyEncDec} and \texttt{GazivEncDec}, we present the reconstruction for subject 3. The reconstructions for all methods are provided by the authors or reported in the original papers. \texttt{SeeligerDCGAN} uses the average of the stimuli representations for the three subjects.}
\label{fig:vis_comp}
\end{figure*}

Figure \ref{fig:vis_comp} A shows the reconstructions of sample stimuli images from the test set, corresponding to subject 1 from \texttt{DIR} dataset. 
%Figure \ref{fig:vis_comp} A, B, and C shows the reconstructions of sample stimuli images from the test set, corresponding to \texttt{DIR}, \texttt{GOD}, and \texttt{vim-1} datasets, respectively. 
The reconstructions from all methods show a close resemblance to the original images in terms of the object shape and position. GAN-based methods, i.e., \linebreak \texttt{ShenDNN+DGN} and \texttt{ShenGAN}, produce sharper and smoother-looking images but in some cases render semantic details absent in the original stimuli (which is confirmed by lower pixelwise MSE and PCC scores). Reconstructions by \texttt{FangSSGAN} are also natural looking and close to real images in terms of shape and color. This is attributed to using a generator network conditioned on both shape and semantic information, which preserves low-level features, such as texture or shape. Reconstruction by nonGAN\footnote{By ``nonGAN'' methods, we mean the models that do not take advantage of the GAN training procedure.} \texttt{BeliyEncDec} are blurry but accurately capture the shape of the stimuli objects. %In terms of color, all methods preserve the color of the stimuli to some extent, except for end-to-end \texttt{ShenGAN}, which according to the authors, preserves color for red-colored shapes only \citep{shen_end--end_2019}.

In addition, we present the reconstructions for \texttt{GOD} dataset in Figure \ref{fig:vis_comp} B. %and for \texttt{vim-1} dataset in Supplementary Figure \ref{fig:vis_comp_vim}. 
Similar to \texttt{DIR} dataset, the GAN-based methods \texttt{MozafariBigBiGAN} and \texttt{RenD-VAE/GAN} produce the most natural-looking images. Visually, \texttt{MozafariBigBiGAN} outperforms other methods in terms of naturalness. However, this comes at the cost of rendering object categories and details different from those presented in the original stimuli. We identified  \texttt{GazivEncDec} and \texttt{RenD-VAE/GAN} as performing relatively better on the reconstruction of shape and color. \texttt{GazivEncDec} is superior in reconstructing high-level details of the image, including shape and background. \texttt{RenD-VAE/GAN} visually outperforms other methods for the reconstruction of color, background, and lower-level details. 
For \texttt{GazivEncDec}, a significant improvement in the reconstruction accuracy was achieved owing to the introduced perceptual similarity loss. According to \cite{ren_reconstructing_2021}, the key factors boosting the reconstruction quality of the \texttt{RenD-VAE/GAN} include the VAE-GAN architecture instead of the standard conditional GAN and visual-feature guidance implemented via GAN-based knowledge distillation.
In \texttt{SeeligerDCGAN} and \texttt{BeliyEncDec}, the reconstructions are blurry, which could be due to the use of pixelwise MSE loss \citep{seeliger_generative_2018}. 

\begin{figure}[h]
\begin{center}
    \includegraphics[width=10cm]{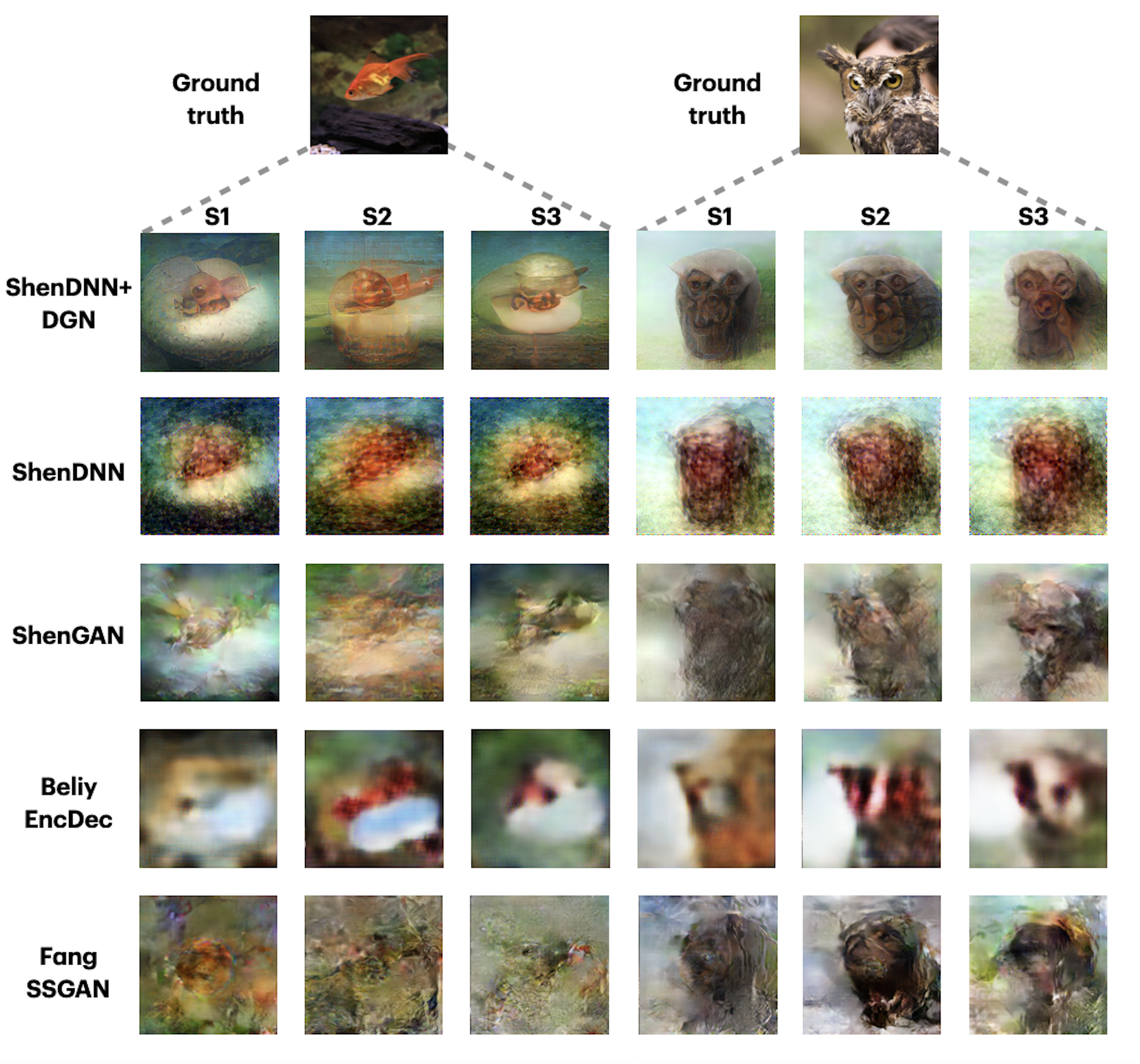}
\end{center}
\caption{
Reconstructions for two images across three subjects from \texttt{DIR} dataset.}
\label{fig:vis_comp_3_subjects}
\end{figure}

Since the \texttt{DIR} dataset comprises three-subject data, we additionally show the reconstructions across the methods corresponding to three different subjects in Figure \ref{fig:vis_comp_3_subjects}. The reconstructions  are shown for the two natural image stimuli. Depending on the subject, the reconstructions by different methods show varying degrees of resemblance to the original stimuli.  For example, the reconstructions from \linebreak \texttt{ShenDNN+DGN}, \texttt{ShenGAN}, and \texttt{BeliyEncDec} are visually better for subject 1, whereas, in reconstructions by other methods, neither color nor shape was preserved. This shows that the selection of a subject in reporting results can lead to a biased evaluation. %Therefore, we perform a quantitative comparison across all the subjects in the dataset in the following Section.

\subsection{Quantitative comparison results on natural images from \texttt{DIR}} \label{sec:quant_comp}

\begin{figure*}[t]
\begin{center}
\includegraphics[width=\linewidth]{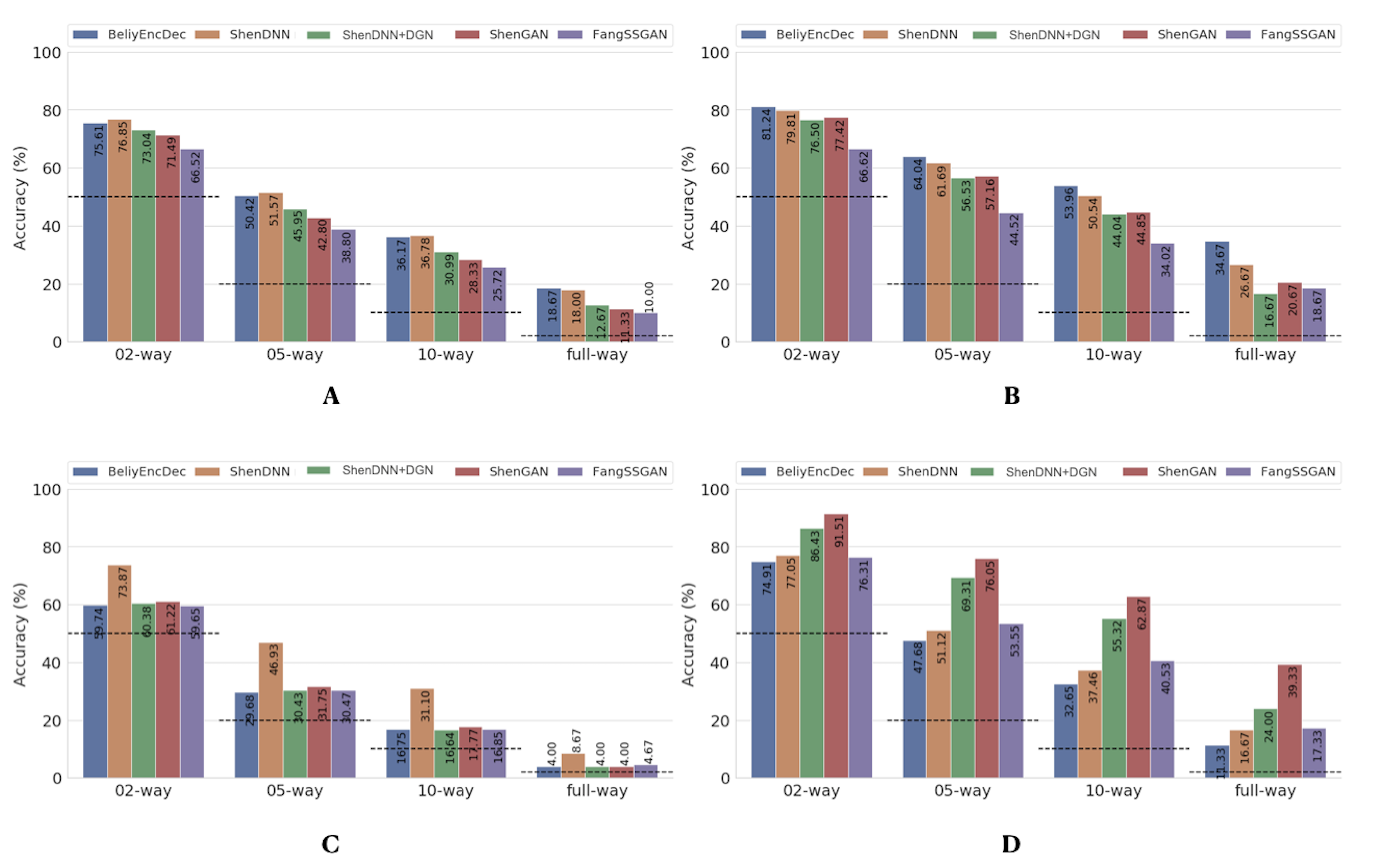}
\end{center}
\caption{Average $n$-way accuracy results computed across subjects using \textbf{(A)} MSE, \textbf{(B)} PCC, \textbf{(C)} SSIM, and \textbf{(D)} PSM metrics on natural images from the \texttt{DIR}. The horizontal dashed lines indicate the chance level for each metric. The full-way comparison corresponds to using all the images in the test set, that is, 50 natural images.}
\label{fig:n_way_images}
\end{figure*}

To eliminate the bias of selecting a specific subject for evaluation, we present both subject-specific and cross-subject average results across multiple metrics on natural images from \texttt{DIR}. % and the individual results per each subject. 
%Subject-specific results are provided in Supplementary Table \ref{}. 
%In this section, we present cross-subject average results across multiple metrics on natural images from \texttt{DIR}. \linebreak 
For comprehensive evaluation, we use three comparison settings described in Section \ref{sec:mertics}: 1) one-to-one comparison; 2) pairwise comparison; and 3) $n$-way comparison. The pairwise evaluation results for natural images across the metrics are shown in Table \ref{table:comp_eval_images}. The $n$-way scores for natural images are presented in Figure \ref{fig:n_way_images}. We find that one-to-one results are not well suited for cross-method comparison. We therefore present a one-to-one comparison in Supplementary Material. % \ref{table:comp_eval_one_to_one}.
%MSE, PCC, and SSIM are categorized as low-level metrics because they operate on a low-level pixel scale. On the other hand, the PSM is used as a high-level metric because it compares image features on multiple levels \textcolor{red}{is this repeated?}. 
The quantitative evaluation of methods is presented based on low-level MSE, PCC, and SSIM metrics first, followed by a comparison using a high-level PSM metric. %The performances of different methods vary with the choice of the subject (see Tables \ref{table:comp_eval_images}, \ref{table:comp_eval_letters}, and \ref{table:comp_eval_shapes}). 

\begin{table*}[h!]
\centering
%\arrayrulecolor{black}
\caption{Pairwise evaluation across the methods on natural images from the \texttt{DIR} dataset. The best results are presented in \textbf{bold}. %The methods that scored the best on $\geq 2$ metrics are shown in \textbf{bold}. 
↑ indicates the higher the better.}
\label{table:comp_eval_images}
\resizebox{12cm}{!}{
\begin{tabular}{clrrrr}
\hline
%\multicolumn{1}{l}{}         &                                           & \multicolumn{4}{c}{\textbf{Pairwise}}                                                           \\ \hline
\multirow{2}{*}{\textbf{Subject}} &
  \multicolumn{1}{c}{\multirow{2}{*}{\textbf{Method}}} &
  \multicolumn{1}{c}{\multirow{2}{*}{\textbf{MSE ↑}}} &
  \multicolumn{1}{c}{\multirow{2}{*}{\textbf{PCC↑}}} &
  \multicolumn{1}{c}{\multirow{2}{*}{\textbf{SSIM ↑}}} &
  \multicolumn{1}{c}{\multirow{2}{*}{\textbf{PSM ↑}}} \\
                             & \multicolumn{1}{c}{}                      & \multicolumn{1}{c}{} & \multicolumn{1}{c}{}    & \multicolumn{1}{c}{} & \multicolumn{1}{c}{}    \\ \hline
\multirow{5}{*}{\textbf{S1}} & ShenDNN              & 75.80                & 80.69                   & \textbf{75.59}       & 77.67                   \\ \cline{2-6} 
                             & ShenDNN+DGN          & 74.53                & 78.98                   & 61.27                & 86.61                   \\ \cline{2-6} 
                             & ShenGAN          & 71.67                & 79.06                   & 62.08                & \textbf{92.33}          \\ \cline{2-6} 
                             & BeliyEncDec       & \textbf{76.94}       & \textbf{86.08}          & 59.67                & 73.14                   \\ \cline{2-6} 
                             & FangSSGAN  & 67.71                & 67.18                   & 60.37                & 76.12                   \\ \hline
\multirow{5}{*}{\textbf{S2}} & ShenDNN             & \textbf{74.98}       & \textbf{77.27}          & \textbf{70.82}       & 77.14                   \\ \cline{2-6} 
                             & ShenDNN+DGN          & 70.78                & 75.43                   & 59.55                & 86.41                   \\ \cline{2-6} 
                             & ShenGAN          & 68.65                & 74.20                   & 59.51                & \textbf{90.41}          \\ \cline{2-6} 
                             & BeliyEncDec       & 71.18                & 76.20                   & 58.94                & 75.22                   \\ \cline{2-6} 
                             & FangSSGAN 
                             & 64.00                & 66.24                   & 58.69                & \textbf{73.71}          \\ \hline
\multirow{5}{*}{\textbf{S3}} & ShenDNN              & \textbf{79.71}       & \textbf{81.47}          & \textbf{75.06}       & 76.20                   \\ \cline{2-6} 
                             & ShenDNN+DGN          & 73.59                & 75.02                   & 60.24                & 86.20                   \\ \cline{2-6} 
                             & ShenGAN          & 74.12                & 78.98                   & 62.08                & \textbf{91.88}          \\ \cline{2-6} 
                             & BeliyEncDec       & 78.61                & \textbf{81.47}          & 60.53                & 76.20                   \\ \cline{2-6} 
                             & FangSSGAN
                             & 67.88                & 66.45                   & 59.96                & 79.02                   \\ \hline
\multirow{5}{*}{\textbf{Average result}} &
  ShenDNN  &
  \textbf{76.83$\pm$2.53} &
  79.81$\pm$2.24 &
  \textbf{73.82$\pm$2.62} &
  77.01$\pm$0.74 \\ \cline{2-6} 
                             & ShenDNN+DGN          & 72.97$\pm$1.95       & 76.48$\pm$2.18          & 60.35$\pm$0.86       & 86.41$\pm$0.20          \\ \cline{2-6} 
                             & ShenGAN         & 71.48$\pm$2.74       & 77.41$\pm$2.78          & 61.22$\pm$1.48       & \textbf{91.54$\pm$1.00} \\ \cline{2-6} 
                             & BeliyEncDec       & 75.58$\pm$3.90       & \textbf{81.25$\pm$4.94} & 59.71$\pm$0.80       & 74.86$\pm$1.56          \\ \cline{2-6} 
                             & FangSSGAN 
                             & 66.53$\pm$2.19       & 66.63$\pm$0.49          & 59.67$\pm$0.87       & 76.29$\pm$2.66          \\ \hline
\multicolumn{1}{l}{}         &                                           & \multicolumn{1}{l}{} & \multicolumn{1}{l}{}    & \multicolumn{1}{l}{} & \multicolumn{1}{l}{}    \\ 
\end{tabular}
}
%\arrayrulecolor{black}
\end{table*}

\textbf{Performance using low-level metrics}. Based on the average results across the subjects shown in Table \ref{table:comp_eval_images} and Figure \ref{fig:n_way_images} A, B, C, two nonGAN methods lead on low-level metrics, namely \texttt{ShenDNN} and \texttt{BeliyEncDec}. Together, they outperform other baselines across three low-level pairwise metrics (i.e., pairwise MSE, pairwise PCC, and pairwise SSIM) as well as across $n$-way MSE and PPC metrics. The high performance of \texttt{BeliyEncDec} on low-level metrics can be attributed to efficient low-level feature extraction via encoder--decoder architecture and to the self-supervised training procedure with the extended set of unlabeled images and fMRI data. 
The high performance of \texttt{ShenDNN} on low-level metrics is potentially due to iterative pixel-level optimization of the reconstructed image.

\textbf{Performance using the high-level PSM metric}. Additionally, we compare the selected methods on the PSM implemented using AlexNet. %, as described in Section \ref{sec:psm}. 
From Table \ref{table:comp_eval_images} and Figure \ref{fig:n_way_images} D, we can see that \texttt{ShenGAN} performs the best on the high-level PSM metric, computed in a pairwise, and $n$-way manner across the subjects and on averages. Overall, GAN-based methods, including \texttt{ShenGAN}, \texttt{ShenDNN+DGN}, and \linebreak \texttt{FangSSGAN}, which were reported to produce more natural-looking images, achieved the top three average results in most cases. This supports the motivation to utilize PSM for measuring high-level visual similarity in images, especially for GAN-based methods whose strength lies in reconstructing high-level visual features and more natural-looking images. We attribute the improved performance of the three methods to using a pretrained generator network and the superior performance of \texttt{ShenGAN} and \texttt{ShenDNN+DGN} to the use of multilayer DNN features for computing the multi-level feature loss. Notably, the performance of all metrics reduces as the $n$-way comparison becomes increasingly harder with an increasing number of samples being used in the comparison.

\section{Discussion}\label{sec:discussion}

Even with a relatively small number of the available open-source reconstruction frameworks, the visual and quantitative results presented in this work can give a general idea of which architectural solution, benchmark dataset, or evaluation framework can be chosen for experimental purposes. %We observed that several factors should be considered to achieve high-quality reconstruction, including network architecture, choice of a loss function, availability of additional data for training, and pretrained network components \textcolor{red}{[too broad and not concrete insight]}. 

Depending on the target of the reconstruction task, it is vital to consider the trade-off between the ``naturalness'' and the fidelity of the reconstruction. 
Generative methods rely on GAN or VAE-GAN-based architectures to produce the most natural-looking images and correspondingly higher PSM scores. However, they often require either external data for training or the use of pretrained network components. The availability of external image datasets for training becomes a significant factor for generating high-quality images for GAN. Most importantly, the methods that perform best at ``naturalness'' do not guarantee that the object categories of reconstruction will always match those of the original images, as in the case of MozafariBigBiGAN. Other non-generative methods developed for natural image reconstruction, such as BeliyEncDec or ShenDNN, do not produce realistic-looking images. However, whenever the fidelity of the reconstructions is preferable, these non-generative methods should be considered, as they exhibit closer similarity to the original images in terms of low-level features, which are supported both visually and quantitatively. %While the introduction of GANs helps to improve the reconstruction quality in some cases, an interesting research question would be how these generative models correspond to the way the brain learns a generative model from the environment.

In this work we advocate the fairness in reconstruction evaluation procedure and discuss several criteria which should be standardized across the methods. At the same time, we believe that the evaluation procedure presented in this work can be further improved in the following ways. %"and we expect the metrics to evolve in several ways". %Next, we will introduce some of the promising research directions that might further advance the reconstruction methods using brain imaging data by improving the quality of methods and adopting more challenging tasks.

% size
%The application of deep learning models has drastically advanced the task of natural image reconstruction. 
% large-scale imaging data + new types of data
\textbf{Availability of large-scale imaging data.}
The primary challenge for current deep learning-based methods is that 
they are required to resolve the limitation of small-size fMRI data. %The first problem, which involves handling noisy data with a low signal-to-noise ratio, is addressed by introducing a feature extraction or encoding framework, which can extract highly informative lower-dimensional representations of the input \textcolor{red}{and by using more number of repeats \citep{beliy_voxels_2019}}. 
Nowadays, the lack of training data is compensated by pretraining DNN components on external image data \citep{shen_end--end_2019, shen_deep_2019, beliy_voxels_2019}, self-supervision on additional image-fMRI pairs \citep{beliy_voxels_2019, gaziv_self-supervised_2020} and generation of new surrogate fMRI via pretrained encoding models \citep{st-yves_generative_2018}. %adopting the available pretrained network components, or data augmentation by generating new plausible data samples.
Several brain imaging datasets are available for reconstruction tasks. However, larger scale %and more challenging 
datasets are still required. The availability of large-scale imaging data may improve current state-of-the-art results and foster research on reconstructing more complex and dynamic visual perception, including imagined images or videos. This, in turn, may lead to broader adoption of the proposed frameworks for real-world purposes.
%+ "By adding qualitatively new types of data." : reconstruction from neural spikes

\textbf{Developing new computational similarity metrics corresponding to human vision.}
% correspondence of these computer-based metrics to human vision
%"By developing new"  similarity metrics". "By developing brain scores that are tuned separately for the non-human primate and the human."
While some of the deep learning methods achieve encouraging results on high-level perceptual similarity metrics, an open question about the correspondence of these computer-based metrics to human vision remains. Because most accuracy evaluation metrics are oriented toward computer vision tasks, they often fail to capture the characteristics of human vision. Research in this direction might further advance natural image reconstruction by developing more advanced learning and evaluation metrics.
%Subjective evaluation involving human feedback may be expensive but remains a reliable metric, serving as a reference for comparing the nonsubjective evaluation results. For human judgment, we advocate performing quantitative subjective evaluations across the baselines rather than presenting results for a proposed method.

%Another potential research direction is to investigate the possibility of transfer learning by training models on a larger natural image dataset and fine-tuning on a smaller dataset with the related target application. For example, models trained to reconstruct natural images can be applied to reconstruct simpler shapes. This is particularly useful for practical applications in the brain image domain, where acquiring new data samples is challenging owing to human subjects' involvement.
\section{Conclusion} \label{sec:conlusion}
This paper presented an overview of state-of-the-art methods for natural image reconstruction task using deep learning. These methods were compared on multiple scales, including architectural design, benchmark datasets, and evaluation metrics. We highlighted several ambiguities with the existing evaluation and presented a standardized empirical assessment of the methods. This evaluation procedure can help researchers in performing a more comprehensive comparative analysis and elucidating the reason for the effectiveness of their method. We hope this study will serve as a foundation for future research on natural image reconstruction targeting fair and rigorous comparisons.

\section*{Author Contributions}

ZR: Conceptualization, Methodology, Software, Writing, Evaluation. QJ: Software, Evaluation, Writing - review and editing. XL: Conceptualization, Data curation, Writing - Original draft preparation. TM: Supervision, Writing - review \& editing. 
%The Author Contributions section is mandatory for all articles, including articles by sole authors. If an appropriate statement is not provided on submission, a standard one will be inserted during the production process. The Author Contributions statement must describe the contributions of individual authors referred to by their initials and, in doing so, all authors agree to be accountable for the content of the work. Please see  \href{http://home.frontiersin.org/about/author-guidelines#AuthorandContributors}{here} for full authorship criteria.

\section*{Funding}
This work was partly supported by JST CREST (Grant Number JPMJCR1687), JSPS Grant-in-Aid for Scientific Research (Grant Number 21K12042, 17H01785), and the New Energy and Industrial Technology Development Organization (Grant Number JPNP20006).

\section*{Acknowledgments}
We thank Professor Yukiyasu Kamitani from Neuroinformatics Lab at Kyoto University for providing valuable comments that improved the manuscript.
We also thank Katja Seeliger, Milad Mozafari, Guy Gaziv, Roman Beliy, and Tao Fang for sharing their reconstructed images and evaluation codes with us.

%\section*{Supplemental Data}
% \href{http://home.frontiersin.org/about/author-guidelines#SupplementaryMaterial}{Supplementary Material} should be uploaded separately on submission, if there are Supplementary Figures, please include the caption in the same file as the figure. LaTeX Supplementary Material templates can be found in the Frontiers LaTeX folder.

%\section*{Data Availability Statement}
%The datasets [GENERATED/ANALYZED] for this study can be found in the [NAME OF REPOSITORY] [LINK].
% Please see the availability of data guidelines for more information, at https://www.frontiersin.org/about/author-guidelines#AvailabilityofData

\bibliographystyle{frontiersinSCNS_ENG_HUMS} % for Science, Engineering and Humanities and Social Sciences articles, for Humanities and Social Sciences articles please include page numbers in the in-text citations
\bibliography{references}

%%% Use this if adding the figures directly in the mansucript, if so, please remember to also upload the files when submitting your article
%%% There is no need for adding the file termination, as long as you indicate where the file is saved. In the examples below the files (logo1.eps and logos.eps) are in the Frontiers LaTeX folder
%%% If using *.tif files convert them to .jpg or .png
%%%  NB logo1.eps is required in the path in order to correctly compile front page header %%%

%%% If you are submitting a figure with subfigures please combine these into one image file with part labels integrated.

\end{document}

\onecolumn
\firstpage{1}

\title[Supplementary Material]{{\helveticaitalic{Supplementary Material}}}

\maketitle

%\section{Supplementary information}

\section{Overview of deep learning methods} \label{sec:DL}

\begin{figure*}[]
\centering
\includegraphics[width=16cm]{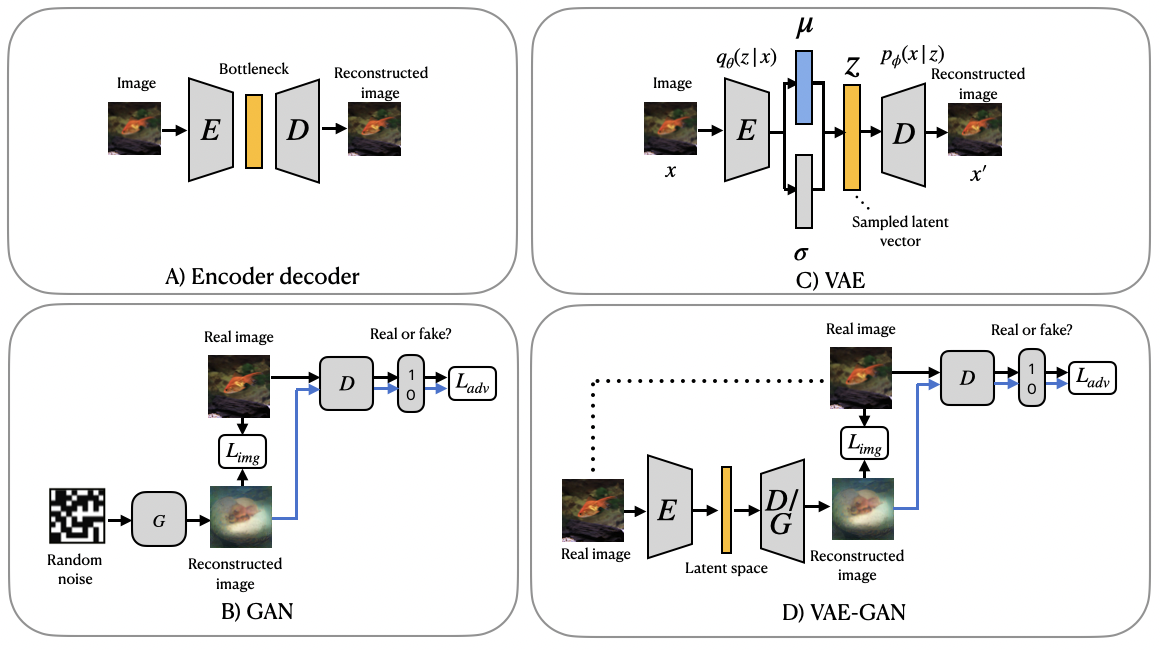}
\caption{\textbf{(A)} Encoder--decoder architecture. \textbf{(B)} VAE. \textbf{(C)} GAN. \textbf{(D)} VAE-GAN.}
\label{fig:dnn_frameworks}
\end{figure*}

\textbf{Encoder--decoder methods.} \label{sec:enc-dec}
%\textcolor{red}{Too broad and non-specific. Add description of separate encoder and decoder, including the features extracted to match ShenDNN.}
In computer vision, encoder--decoder models are multilayer models that incorporate convolutional layers. Encoder--decoder models are widely used in image-to-image translation \citep{isola_image--image_2017} and sequence-to-sequence models \citep{cho_learning_2014}. They learn the mapping from an input domain to an output domain via a two-stage architecture: an encoder ${E}$ that compresses the input to the latent space representation $\mathbf{z} = {E}(\mathbf{x})$ and a decoder ${{D}}$ that produces the output from the latent representation $\mathbf{y}={D}(\mathbf{z})$ (see Figure \ref{fig:dnn_frameworks} A) \citep{minaee_image_2021}. %The compressed latent representation vector $z$ serves as a bottleneck, which encodes a low-dimensional representation of the input $x$. 
The model is trained to minimize the reconstruction error, which is the difference between the reconstructed image and ground-truth image.

\textbf{Generative Adversarial Network (GAN). }
A GAN, shown in Figure \ref{fig:dnn_frameworks} B, 
contains generator and discriminator networks. In the image generation task, the generator ${{G}}$ takes a random noise vector $\mathbf{z}$ (generally sampled from a Gaussian distribution) and generates a fake sample $G(\mathbf{z})$ with the same statistics as the training set images. The task of the discriminator network ${{D}}$ is to distinguish the generated fake sample $G(\mathbf{z})$ from the real sample $\mathbf{x}$ by maximizing the probability $D(\mathbf{x})$ and minimizing $D(G(\mathbf{z}))$. This learning process is formulated as a zero-sum game with the following mini--max loss \citep{goodfellow_generative_2014}:
\begin{equation}
    E_{\mathbf{x}}[\log (D(\mathbf{x}))]+E_{\mathbf{z}}[\log (1-D(G(\mathbf{z})))]
\end{equation}
where $E_{\mathbf{x}}$ and $E_{\mathbf{z}}$ are the expected values of overall real and generated fake samples, $G(\mathbf{z})$. During training, the generator's ability to generate realistic images continually improves until the discriminator is unable to distinguish the difference between a real sample and a generated fake one. 
GAN-based frameworks have several desirable properties compared to other generative methods. First, GANs do not require strong assumptions regarding the form of the output probability distribution. Second, adversarial training, which uses the discriminator, allows unsupervised training of the GAN \citep{st-yves_generative_2018}.

\textbf{The variational autoencoder (VAE)} proposed by \cite{kingma_auto-encoding_2014} is a popular generative algorithm used in neural decoding (see Figure \ref{fig:dnn_frameworks} C). Similar to autoencoders, the VAE is composed of an encoder and a decoder. The encoder $q_{\phi}({\mathbf{z}}|\mathbf{x})$, parameterized with $\phi$, embeds the input $\mathbf{x}$ into a latent representation $\mathbf{z}$, and decoder $p_{\theta}(\mathbf{x}|\mathbf{z})$, parameterized with $\theta$, maps the latent representation $\mathbf{z}$ back to the observation. But rather than encoding a latent vector, VAE encodes a distribution over the latent space,  which makes the generative process possible. Thus, the goal of VAE is to find a distribution of the latent variable $\mathbf{z} \sim q_{\phi}(\mathbf{z}|\mathbf{x})$,
where $q_{\phi}(\mathbf{z}|\mathbf{x})$ is a Gaussian distribution whose mean and standard deviation are the output of the encoder. The distribution $q_{\phi}(\mathbf{z}|\mathbf{x})$ is regularized to be a simple Gaussian distribution based on the Kullback-Leibler divergence between $q_{\phi}(\mathbf{z}|\mathbf{x})$, and the prior distribution $p(\mathbf{z})$ is  assumed to follow a Gaussian distribution ${\mathcal {N}}(0 , 1)$. Then we can sample $\mathbf{z} \sim q_{\phi}(\mathbf{z}|\mathbf{x})$ to generate new samples $\mathbf{x}' \sim  p_{\theta}(\mathbf{x}|\mathbf{z})$. The learned latent representation $q_{\phi}(\mathbf{z}|\mathbf{x})$ approximates a true posterior distribution $p_{\theta}(\mathbf{z}|\mathbf{x})$, which is normally intractable. VAE loss consists of two terms:
\begin{equation}\label{eq:vae}
 \mathcal{L}(\theta, \phi ; \mathbf{x}, \mathbf{z})=\mathbb{E}_{q_{\phi}(\mathbf{z} \mid \mathbf{x})}\left[\log p_{\theta}(\mathbf{x} \mid \mathbf{z})\right]-D_{K L}\left(q_{\phi}(\mathbf{z} \mid \mathbf{x}) \| p(\mathbf{z})\right).
\end{equation}
The first term maximizes the reconstruction likelihood by penalizing the reconstruction error. The second term is a regularizer that enforces that the learned distribution $q_{\phi}(\mathbf{z}|\mathbf{x})$ is similar to the prior distribution $p(\mathbf{z})$ by minimizing the distance error between two probability distributions measured by the Kullback--Leibler divergence. 

%To ensure that the latent space is regularised, the encoder is trained to return the mean ($\mu$) and the standard deviation ($\sigma$) for each latent variable. Then, the latent vector is randomly sampled from this distribution and passed to the decoder for reconstruction. Due to this random sampling, it is difficult for the errors to propagate through the network. To make gradient backpropagation possible, a reparameterization trick is used during the random sampling process, which represents latent vector ${z}$ as a function of the encoder’s output:
%$z={\mu}+{\sigma} {\epsilon}$, where ${\epsilon}$ is an auxiliary noise variable ${\epsilon} \sim \mathcal{N}(0,1)$, and $\mu$ and $\sigma$ are mean and standard deviation of the approximate posterior. We refer to \cite{kingma_auto-encoding_2014} for details on mathematical derivations for VAE. 

\textbf{VAE-GAN} is a hybrid model by \cite{larsen_autoencoding_2016} integrates both the VAE and GAN. VAE-GAN combines VAE to produce compressed latent features $\mathbf{z}$ and GAN discriminator, which learns to discriminate between fake and real images. The VAE decoder and GAN generator are combined into one. Figure \ref{fig:dnn_frameworks} D illustrates the VAE-GAN framework.
The VAE-GAN loss is computed as follows:
\begin{equation}
\label{eq:vae-gan}
    \mathcal{L}_{VAE-GAN} = \mathcal{L}_{Dis} + D_{K L}\left(q_{\phi}({\mathbf{z}} \mid {\mathbf{x}}) \| p({\mathbf{z}})\right) + \mathcal{L}_{GAN}.
\end{equation}
Compared to the VAE loss in equation \ref{eq:vae}, VAE-GAN loss includes $\mathcal{L}_{GAN}$ loss, and the reconstruction loss $\mathcal{L}(\mathbf{x}, \mathbf{x'})$ is replaced with a reconstruction error expressed using the intermediate features of the GAN discriminator $\mathcal{L}_{Dis}$.

\section{Reconstructed image resolutions}
Reconstructed image resolutions reported by surveyed works for natural image datasets are shown in Table \ref{table:comp_resolution}.

\begin{table}
    \centering
    \caption{Reconstructed image resolutions reported by surveyed works for natural image datasets.}
    \label{table:comp_resolution}
\resizebox{10cm}{!}{
\begin{tabular}{lcc}
\hline
\multicolumn{1}{c}{\textbf{Reference}} & \textbf{Dataset} & \textbf{\begin{tabular}[c]{@{}c@{}}Reconstructed image \\ size\end{tabular}} \\ \hline
\cite{seeliger_generative_2018}      & \begin{tabular}[c]{@{}c@{}}vim-1\\ GOD\end{tabular}  & 64 x 64   \\ \hline
\cite{st-yves_generative_2018}       & vim-1                                                & 32 x 32   \\ \hline
\cite{shen_deep_2019}                & DIR                                                  & 224 x 224 \\ \hline
\cite{shen_end--end_2019}            & DIR                                                  & 227 x 227 \\ \hline
\cite{beliy_voxels_2019}       & \begin{tabular}[c]{@{}c@{}}GOD \\ vim-1\end{tabular} & 112 x 112 \\ \hline
\cite{vanrullen_reconstructing_2019} & Faces                                                & 128 x 128 \\ \hline
\cite{gaziv_self-supervised_2020}    & \begin{tabular}[c]{@{}c@{}}GOD \\ vim-1\end{tabular} & 112 x 112 \\ \hline
\cite{qiao_biggan-based_2020}        & vim-1                                                & 128 x 128 \\ \hline
\cite{fang_reconstructing_2020}      & DIR                                                  & 256 x 256 \\ \hline
\cite{mozafari_reconstructing_2020}  & GOD                                                  & 256 x 256 \\ \hline
\cite{ren_reconstructing_2021}       & GOD                                                  & 100 x 100 \\ \hline
\end{tabular}
}
\end{table}

\begin{figure}[h]
\begin{center}
    \includegraphics[width=10cm]{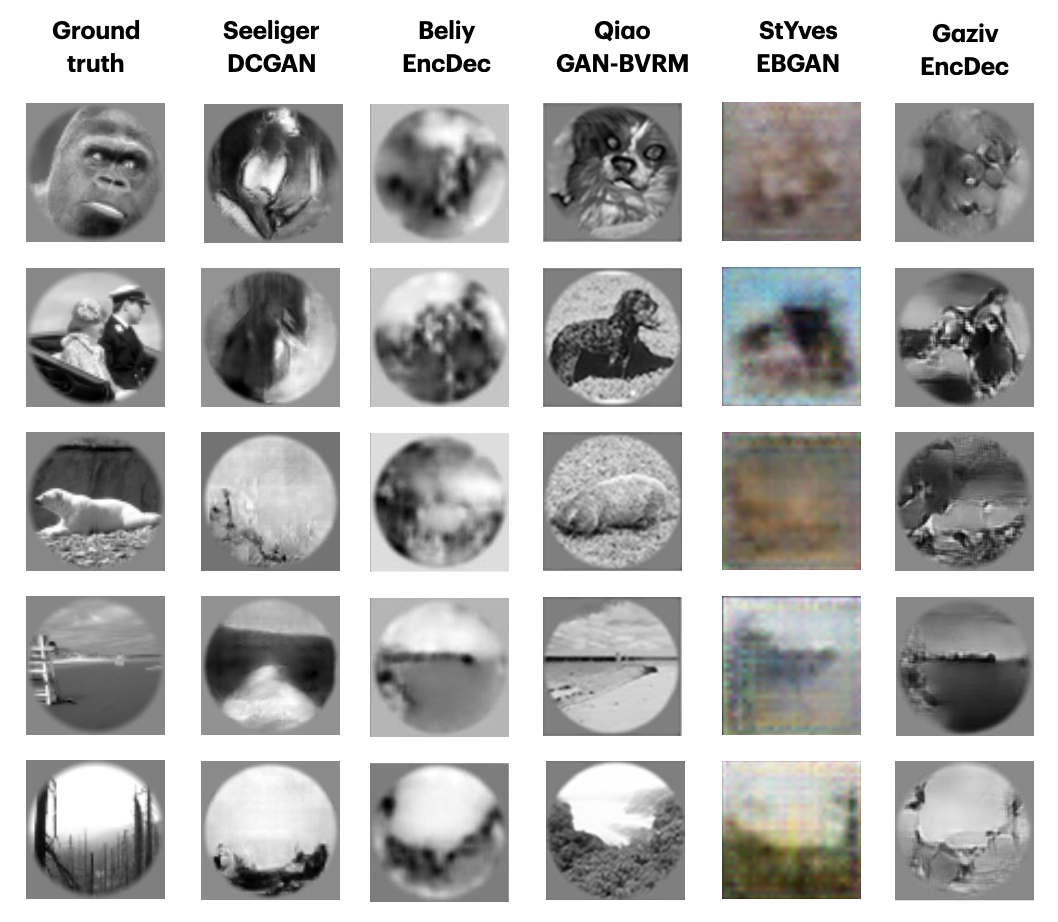}
\end{center}
\caption{
Visual comparison of methods on \texttt{vim-1} dataset for subject 1. The reconstructions for all methods are provided by the authors or reported in the original papers. The reconstructions for \texttt{BeliyEncDec} \citep{beliy_voxels_2019}, \texttt{QiaoGAN-BVRM} \citep{qiao_biggan-based_2020}, and \texttt{StYvesEBGAN} \citep{st-yves_generative_2018} were taken from the original papers. \texttt{SeeligerDCGAN} uses the average of the stimuli representations for the three subjects.}
\label{fig:vis_comp_vim}
\end{figure}

\section{Visual inspection}
For \texttt{vim-1} dataset, the \texttt{SeeligerDCGAN}, \texttt{StYvesEBGAN}, and \texttt{QiaoGAN-BVRM} methods used GAN as natural image prior to generate natural-looking images. However, low-resolution reconstructions of \texttt{StYvesEBGAN} can only capture high-level image features such as the line of the horizon but do not preserve low-level details of the stimuli. 
The high-resolution samples generated from \texttt{SeeligerDCGAN} and \texttt{QiaoGAN-BVRM} are more realistic and capture the high-level attributes of the image. In \texttt{QiaoGAN-BVRM}, the prior decoded categories play a central role in reconstructing high-level semantic content.  
For example, the reconstruction of the image in the first row of Figure \ref{fig:vis_comp_vim} captures the facial features of the object.
However, in several other cases, the reconstructed object categories do not correspond to the true ones, as shown in the second row of Figure \ref{fig:vis_comp_vim} (a reconstruction of ``dog'' object for stimuli containing a ``human''). The authors attribute this to the inconsistency between image categories in \texttt{vim-1} stimuli images and ImageNet, which was used for pretraining the model.
\texttt{BeliyEncDec}, \linebreak \texttt{GazivEncDec}, and  \texttt{StYvesEBGAN} do not clearly reconstruct identifiable objects, although their performance is the best in preserving the shape and global layout.

\section{Comparative one-to-one evaluation on natural images}

Comparative evaluation using one-to-one and pairwise settings across the methods on natural images from the \texttt{DIR} dataset \citep{shen_deep_2019} is shown in Table \ref{table:comp_eval_one_to_one}.

\begin{table*}[]
\centering
%\arrayrulecolor{black}
\caption{One-to-one comparative evaluation across the methods on natural images from the \texttt{DIR} dataset \citep{shen_deep_2019}. The best results are presented in \textbf{bold}. The methods that scored the best on $\geq 2$ metrics are shown in \textbf{bold}. ↓ indicates that the lower the better, and ↑ indicates the higher the better.}
\label{table:comp_eval_one_to_one}
\resizebox{\linewidth}{!}{
\begin{tabular}{clrrrr}
\hline
\multirow{2}{*}{\textbf{Subject}} &
  \multicolumn{1}{c}{\multirow{2}{*}{\textbf{Method}}} &
  \multicolumn{1}{c}{\multirow{2}{*}{\textbf{MSE ↓}}} &
  \multicolumn{1}{c}{\multirow{2}{*}{\textbf{PCC ↑}}} &
  \multicolumn{1}{c}{\multirow{2}{*}{\textbf{SSIM ↑}}} &
  \multicolumn{1}{c}{\multirow{2}{*}{\textbf{PSM ↓}}} \\
 &
  \multicolumn{1}{c}{} &
  \multicolumn{1}{c}{} &
  \multicolumn{1}{c}{} &
  \multicolumn{1}{c}{} &
  \multicolumn{1}{c}{} \\ \hline
\multirow{5}{*}{\textbf{S1}} &
  ShenDNN \citep{shen_deep_2019} &
  0.11$\pm$0.05 &
  0.28$\pm$0.26 &
  0.11$\pm$0.05 &
  0.93$\pm$0.09 \\ \cline{2-6} 
 &
  ShenDNN+DGN \citep{shen_deep_2019} &
  0.10$\pm$0.05 &
  \textbf{0.32$\pm$0.29} &
  0.26$\pm$0.13 &
  0.70$\pm$0.08 \\ \cline{2-6} 
 &
  ShenGAN \citep{shen_end--end_2019} &
  \textbf{0.09$\pm$0.04} &
  0.31$\pm$0.26 &
  0.26$\pm$0.14 &
  \textbf{0.66$\pm$0.06} \\ \cline{2-6} 
 &
  BeliyEncDec \citep{beliy_voxels_2019} &
  \textbf{0.09$\pm$0.04} &
  \textbf{0.32$\pm$0.22} &
  \textbf{0.31$\pm$0.13} &
  0.88$\pm$0.08 \\ \cline{2-6} 
 &
  FangSSGAN \citep{fang_reconstructing_2020} &
  0.10$\pm$0.06 &
  0.21$\pm$0.27 &
  0.22$\pm$0.12 &
  0.70$\pm$0.07 \\ \hline
\multirow{5}{*}{\textbf{S2}} &
  ShenDNN \citep{shen_deep_2019} &
  0.11$\pm$0.04 &
  0.27$\pm$0.25 &
  0.10$\pm$0.04 &
  0.94$\pm$0.09 \\ \cline{2-6} 
 &
  ShenDNN+DGN \citep{shen_deep_2019} &
  0.10$\pm$0.05 &
  \textbf{0.29$\pm$0.29} &
  0.25$\pm$0.12 &
  0.71$\pm$0.08 \\ \cline{2-6} 
 &
  ShenGAN \citep{shen_end--end_2019} &
  \textbf{0.09$\pm$0.04} &
  0.26$\pm$0.24 &
  0.24$\pm$0.14 &
  \textbf{0.68$\pm$0.06} \\ \cline{2-6} 
 &
  BeliyEncDec \citep{beliy_voxels_2019} &
  0.12$\pm$0.05 &
  0.22$\pm$0.24 &
  \textbf{0.30$\pm$0.14} &
  0.87$\pm$0.09 \\ \cline{2-6} 
 &
  FangSSGAN \citep{fang_reconstructing_2020} &
  \textbf{0.09$\pm$0.04} &
  0.22$\pm$0.24 &
  0.19$\pm$0.11 &
  0.70$\pm$0.07 \\ \hline
\multirow{5}{*}{\textbf{S3}} &
  ShenDNN \citep{shen_deep_2019} &
  0.11$\pm$0.05 &
  0.29$\pm$0.25 &
  0.10$\pm$0.04 &
  0.94$\pm$0.08 \\ \cline{2-6} 
 &
  ShenDNN+DGN \citep{shen_deep_2019} &
  \textbf{0.09$\pm$0.06} &
  0.30$\pm$0.30 &
  0.25$\pm$0.13 &
  0.71$\pm$0.08 \\ \cline{2-6} 
 &
  ShenGAN \citep{shen_end--end_2019} &
  \textbf{0.09$\pm$0.05} &
  \textbf{0.31$\pm$0.24} &
  0.25$\pm$0.14 &
  \textbf{0.66$\pm$0.06} \\ \cline{2-6} 
 &
  BeliyEncDec \citep{beliy_voxels_2019} &
  \textbf{0.09$\pm$0.04} &
  \textbf{0.31$\pm$0.24} &
  \textbf{0.31$\pm$0.14} &
  0.88$\pm$0.10 \\ \cline{2-6} 
 &
  FangSSGAN \citep{fang_reconstructing_2020} &
  0.10$\pm$0.05 &
  0.21$\pm$0.25 &
  0.18$\pm$0.10 &
  0.69$\pm$0.06 \\ \hline
\multirow{5}{*}{\textbf{Average result}} &
  ShenDNN \citep{shen_deep_2019} &
  0.11$\pm$0.05 &
  0.28$\pm$0.25 &
  0.11$\pm$0.05 &
  0.94$\pm$0.01 \\ \cline{2-6} 
 &
  ShenDNN+DGN \citep{shen_deep_2019} &
  0.10$\pm$0.05 &
  \textbf{0.30$\pm$0.29} &
  0.25$\pm$0.13 &
  0.71$\pm$0.00 \\ \cline{2-6} 
 &
  ShenGAN \citep{shen_end--end_2019} &
  \textbf{0.09$\pm$0.04} &
  0.29$\pm$0.25 &
  0.25$\pm$0.14 &
  \textbf{0.67$\pm$0.01} \\ \cline{2-6} 
 &
  BeliyEncDec \citep{beliy_voxels_2019} &
  0.10$\pm$0.04 &
  0.28$\pm$0.23 &
  \textbf{0.30$\pm$0.13} &
  0.87$\pm$0.01 \\ \cline{2-6} 
 &
  FangSSGAN \citep{fang_reconstructing_2020} &
  0.10$\pm$0.05 &
  0.21$\pm$0.25 &
  0.20$\pm$0.11 &
  0.70$\pm$0.00 \\ \hline
\end{tabular}
}
%\arrayrulecolor{black}
\end{table*}

\begin{figure*}[]
\begin{center}
\includegraphics[width=\linewidth]{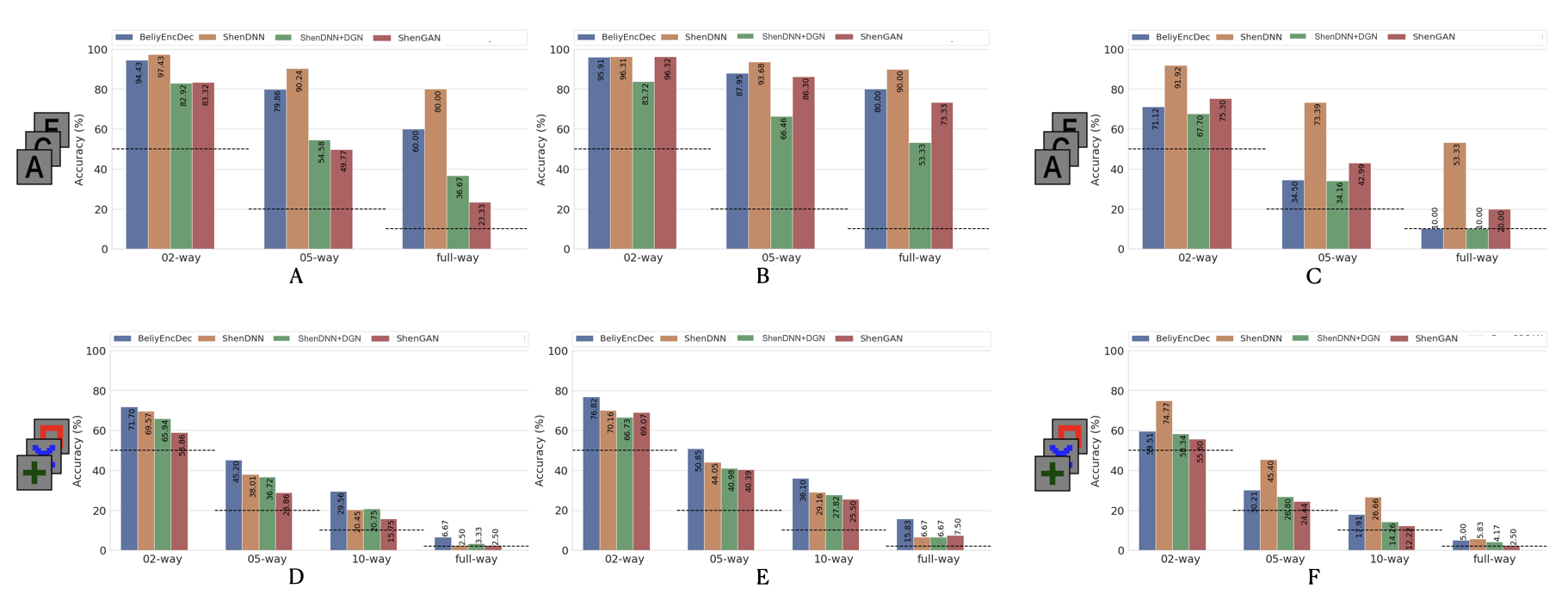}
\end{center}
\caption{Average $n$-way accuracy results computed across subjects using MSE, PCC, and SSIM metrics on different datasets from the \texttt{DIR}: alphabetical letters (first row), and artificial shapes (second row). The horizontal dashed lines indicate the chance level for each metric. The full-way comparison corresponds to using all the images in the test set, that is, 50 natural images, 10 images in alphabetical letters, and 40 images in artificial shapes from the \texttt{DIR}).}
\label{fig:n_way_appendix}
\end{figure*}

\section{Generalization to alphabetical letters and artificial shapes} \label{sec:generalization}

Other than natural images, \texttt{DIR} dataset contains alphabetical letters and artificial shapes images, which were originally designed to be used only for inference by a model pretrained on natural image data \citep{shen_deep_2019, shen_end--end_2019}. 
The ability of the models to generalize to artificial images from different image domains can also indicate that the reconstruction model does not simply match to exemplars \citep{shen_deep_2019}. In particular, similar to \cite{shen_end--end_2019, shen_deep_2019}, we test whether the knowledge acquired from natural image reconstruction from fMRI can be transferred to the task of reconstructing shapes and letters. 
We perform the evaluation on %natural images, 
shapes, and letters from the \texttt{DIR} based on the metrics: MSE, PCC, and SSIM.

%\section{Generalization to artificial shapes and alphabetical letters} \label{sec:generalization}

In addition to running the generalizability experiment for \texttt{ShenDNN+DGN} \citep{shen_deep_2019} and \texttt{ShenGAN} \citep{shen_end--end_2019}, which demonstrated their generalization power, we ran pretrained \linebreak \texttt{BeliyEncDec} \citep{beliy_voxels_2019} on alphabetical letters and artificial shapes images from \texttt{DIR}.  Generalizing the model trained on natural images to different data is challenging but necessary to ensure that the model reconstructs images from fMRI activity rather than memorizing or making matches to the stimuli \citep{shen_deep_2019}.

\begin{figure*}[h]
\centering
\includegraphics[width=15cm]{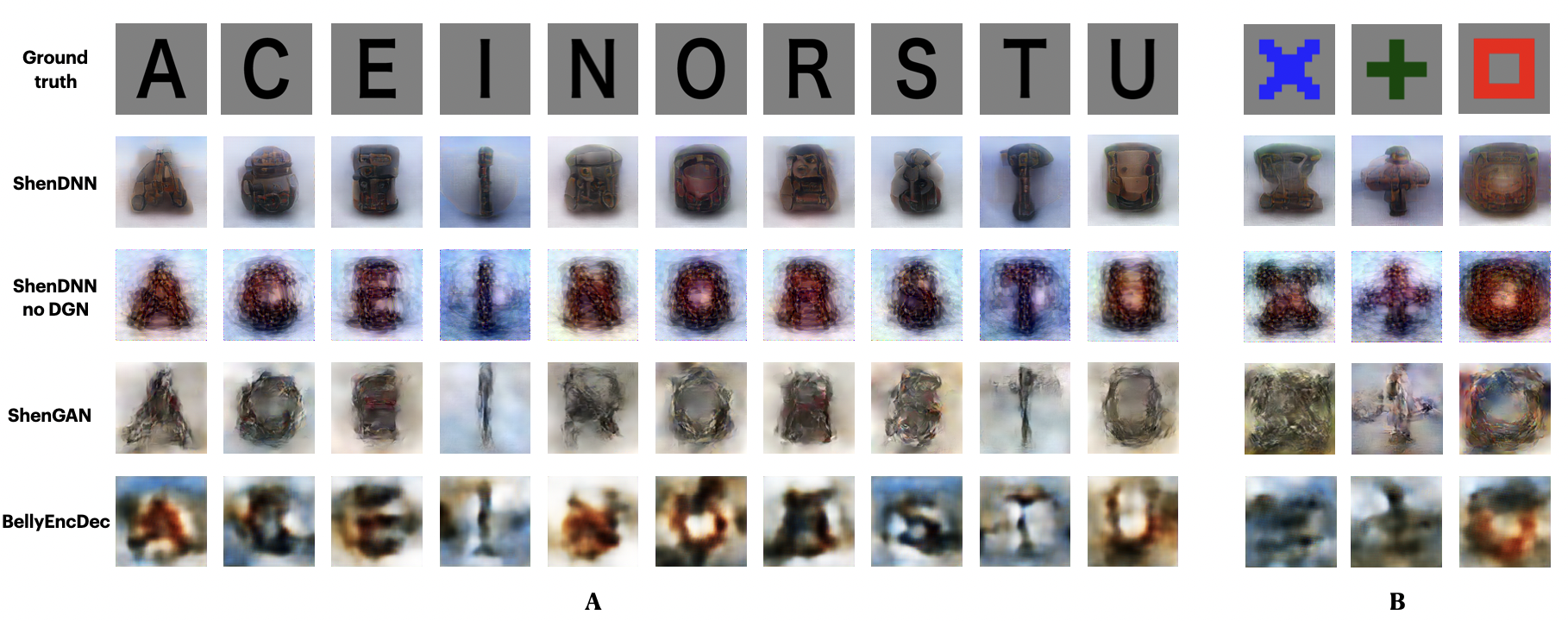}
\caption{Reconstructed samples from generalization task: \textbf{(A)} alphabetical letters and \textbf{(B)} artificial shapes}
\label{fig:recon_letters_shapes}
\end{figure*}

The visual reconstructions of samples from alphabetical letters  and artificial shapes are presented in Figures  \ref{fig:recon_letters_shapes} A and B. For alphabetical letters, \texttt{ShenDNN+DGN} and \texttt{ShenGAN} produce reconstructions with clear contours and backgrounds. For the artificial shapes dataset, the images were successfully reconstructed by all methods, but the reconstruction of red-colored shapes was particularly clear in \texttt{BeliyEncDec}. In reconstructions for both artificial shapes and alphabetical letters, \texttt{BeliyEncDec}  \citep{beliy_voxels_2019} produces background details not present in the original stimuli, as well as greater diversity in colors between images.

The qualitative and quantitative results presented in terms of one-to-one and pairwise MSE, PCC, and SSIM  for alphabetical letters and artificial shapes are shown in Table \ref{table:comp_eval_letters} and Table \ref{table:comp_eval_shapes}. Three methods, \texttt{ShenDNN}, \texttt{ShenGAN}, and \texttt{BeliyEncDec}, perform the best on alphabetical letters images across at least two metrics. Similar to natural images, nonGAN methods, namely \texttt{ShenDNN} and  \texttt{BeliyEncDec}, outperform other methods on low-level metrics. %, and \texttt{ShenGAN} outperforms on the high-level PSM. 
For artificial shapes, \texttt{BeliyEncDec} \citep{beliy_voxels_2019} outperformed other methods on four lower-level metrics. \texttt{ShenDNN} \citep{shen_deep_2019} performed significantly better on pairwise and 2-way SSIM, achieving a score of approximately 74.70\%. 

%The $n$-way PSM results for alphabetical letters and artificial shapes are shown in Figure \ref{fig:n_way_SSIM_PSM} D and F. Comparing the PSM results, we can see that the reconstruction performance of GAN-based methods gradually becomes more on par with nonGAN methods with a decreasing level of perceptual complexity, in other words, as we shift from images to letters to shapes. This emphasizes the importance of using the high-level PSM metric for natural images with higher visual complexity. From Figure \ref{fig:n_way_appendix} A, B, E, and F, showing the $n$-way MSE and PCC results for natural images and artificial shapes, it can be seen that the common trend is that nonGAN methods, that is, \texttt{BeliyEncDec} and \texttt{ShenDNN}, outperformed the other methods. However, as we move from lower-level metrics to the metrics with the gradually increasing perceptual awareness level %---SSIM and PSM are shown in Figure \ref{fig:n_way_SSIM_PSM} A, B, E, F---the trend reverses in favor of GAN-based methods.} 

%Although pairwise PSM is not directly comparable with the pairwise human judgment metric reported by \citep{shen_deep_2019} and \citep{shen_end--end_2019}, the two metrics correspond in terms of higher scores on natural images and alphabetical letters and lower scores on artificial shapes. For example, the human judgment scores reported for \texttt{ShenGAN} by the authors \citep{shen_end--end_2019} were  95.70\%, 96.40\%, and 92.70\% for natural images, alphabetical letters, and artificial shapes, respectively.

Overall, the performance of all $n$-way metrics is better for  alphabetical letters compared to %natural images and
artificial shapes from \texttt{DIR}. The highest scores on alphabetical letters were achieved by \texttt{ShenDNN} on 2-way MSE and 2-way PCC metrics, (97.43\% and 96.31\%, respectively).

\begin{table*}[]
\centering
%\arrayrulecolor{black}
\caption{Comparative evaluation across the methods on alphabetical letters from \texttt{DIR} dataset \citep{shen_deep_2019}. The best results are presented in \textbf{bold}. ↓ indicates that a lower value is better, and ↑ indicates a higher value is better.}
\label{table:comp_eval_letters}
\resizebox{\linewidth}{!}{
\begin{tabular}{clrrrrrr}
\hline
\multicolumn{1}{l}{} &
   &
  \multicolumn{3}{c}{\textbf{One-to-one}} &
  \multicolumn{3}{c}{\textbf{Pairwise}} \\ \hline
\multirow{2}{*}{\textbf{Subject}} &
  \multicolumn{1}{c}{\multirow{2}{*}{\textbf{Method}}} &
  \multicolumn{1}{c}{\multirow{2}{*}{\textbf{MSE ↓}}} &
  \multicolumn{1}{c}{\multirow{2}{*}{\textbf{PCC ↑}}} &
  \multicolumn{1}{c}{\multirow{2}{*}{\textbf{SSIM ↑}}} &
  \multicolumn{1}{c}{\multirow{2}{*}{\textbf{MSE ↑}}} &
  \multicolumn{1}{c}{\multirow{2}{*}{\textbf{PCC ↑}}} &
  \multicolumn{1}{c}{\multirow{2}{*}{\textbf{SSIM ↑}}} \\
 &
  \multicolumn{1}{c}{} &
  \multicolumn{1}{c}{} &
  \multicolumn{1}{c}{} &
  \multicolumn{1}{c}{} &
  \multicolumn{1}{c}{} &
  \multicolumn{1}{c}{} &
  \multicolumn{1}{c}{} \\ \hline
\multirow{4}{*}{\textbf{S1}} &
  ShenDNN \citep{shen_deep_2019} &
  0.08$\pm$0.01 &
  0.51$\pm$0.06 &
  0.17 $\pm$ 0.01 &
  \textbf{98.89} &
  \textbf{100.00} &
  \textbf{96.67} \\ \cline{2-8} 
 &
  ShenDNN+DGN \citep{shen_deep_2019} &
  0.07$\pm$0.01 &
  \textbf{0.53$\pm$0.04} &
  0.5 $\pm$ 0.04 &
  90.00 &
  91.11 &
  73.33 \\ \cline{2-8} 
 &
  ShenGAN \citep{shen_end--end_2019} &
  0.07$\pm$0.01 &
  0.52$\pm$0.06 &
  \textbf{0.54 $\pm$ 0.06} &
  83.33 &
  94.44 &
  74.44 \\ \cline{2-8} 
 &
  BeliyEncDec \citep{beliy_voxels_2019} &
  \textbf{0.06$\pm$0.02} &
  0.41$\pm$0.08 &
  \textbf{0.54 $\pm$ 0.06} &
  96.67 &
  \textbf{100.00} &
  76.67 \\ \hline
\multirow{4}{*}{\textbf{S2}} &
  ShenDNN \citep{shen_deep_2019} &
  0.08$\pm$0.01 &
  0.44$\pm$0.06 &
  0.14 $\pm$ 0.02 &
  \textbf{95.56} &
  88.89 &
  \textbf{87.78} \\ \cline{2-8} 
 &
  ShenDNN+DGN \citep{shen_deep_2019} &
  0.07$\pm$0.01 &
  \textbf{0.46$\pm$0.06} &
  0.46 $\pm$ 0.03 &
  73.33 &
  67.78 &
  64.44 \\ \cline{2-8} 
 &
  ShenGAN \citep{shen_end--end_2019} &
  0.07$\pm$0.01 &
  0.37$\pm$0.14 &
  0.51 $\pm$ 0.08 &
  83.33 &
  \textbf{96.67} &
  68.89 \\ \cline{2-8} 
 &
  BeliyEncDec \citep{beliy_voxels_2019} &
  \textbf{0.05$\pm$0.01} &
  0.35$\pm$0.08 &
  \textbf{0.53 $\pm$ 0.05} &
  90.00 &
  90.00 &
  60.00 \\ \hline
\multirow{4}{*}{\textbf{S3}} &
  ShenDNN \citep{shen_deep_2019} &
  0.09$\pm$0.01 &
  0.51$\pm$0.04 &
  0.14 $\pm$ 0.01 &
  \textbf{97.78} &
  \textbf{100.00} &
  \textbf{91.11} \\ \cline{2-8} 
 &
  ShenDNN+DGN \citep{shen_deep_2019} &
  0.06$\pm$0.01 &
  \textbf{0.52$\pm$0.05} &
  0.46 $\pm$ 0.04 &
  85.56 &
  92.22 &
  65.56 \\ \cline{2-8} 
 &
  ShenGAN \citep{shen_end--end_2019} &
  0.08$\pm$0.02 &
  0.49$\pm$0.13 &
  0.53 $\pm$ 0.06 &
  83.33 &
  97.78 &
  82.22 \\ \cline{2-8} 
 &
  BeliyEncDec \citep{beliy_voxels_2019} &
  \textbf{0.05$\pm$0.02} &
  0.35$\pm$0.12 &
  \textbf{0.55 $\pm$ 0.04} &
  96.67 &
  97.78 &
  76.67 \\ \hline
\multirow{4}{*}{\textbf{Average result}} &
  ShenDNN \citep{shen_deep_2019} &
  0.08$\pm$0.00 &
  0.49$\pm$0.04 &
  0.15 $\pm$ 0.02 &
  \textbf{97.41$\pm$1.70} &
  \textbf{96.30$\pm$6.41} &
  \textbf{91.85$\pm$4.49} \\ \cline{2-8} 
 &
  ShenDNN+DGN \citep{shen_deep_2019} &
  0.07$\pm$0.00 &
  \textbf{0.51$\pm$0.04} &
  0.47 $\pm$ 0.02 &
  82.96$\pm$8.63 &
  83.70$\pm$13.80 &
  67.78$\pm$4.84 \\ \cline{2-8} 
 &
  ShenGAN \citep{shen_end--end_2019} &
  0.07$\pm$0.01 &
  0.46$\pm$0.08 &
  0.53 $\pm$ 0.02 &
  83.33$\pm$0.00 &
  \textbf{96.30$\pm$1.70} &
  75.19$\pm$6.70 \\ \cline{2-8} 
 &
  BeliyEncDec \citep{beliy_voxels_2019} &
  \textbf{0.06$\pm$0.01} &
  0.37$\pm$0.03 &
  \textbf{0.54 $\pm$ 0.01} &
  94.44$\pm$3.85 &
  95.93$\pm$5.25 &
  71.11$\pm$9.62 \\ \hline
\end{tabular}
}
%\arrayrulecolor{black}
\end{table*}

\begin{table*}[]
\centering
%\arrayrulecolor{black}
\caption{Comparative evaluation across the methods on artificial shapes from \texttt{DIR} dataset \citep{shen_deep_2019}. The best results are presented in \textbf{bold}. ↓ indicates that the lower the better, and ↑ indicates the higher the better.}
\label{table:comp_eval_shapes}
\resizebox{\linewidth}{!}{
\begin{tabular}{clrrrrrr}
\hline
\multicolumn{1}{l}{} &
   &
  \multicolumn{3}{c}{\textbf{One-to-one}} &
  \multicolumn{3}{c}{\textbf{Pairwise}} \\ \hline
\multirow{2}{*}{\textbf{Subject}} &
  \multicolumn{1}{c}{\multirow{2}{*}{\textbf{Method}}} &
  \multicolumn{1}{c}{\multirow{2}{*}{\textbf{MSE ↓}}} &
  \multicolumn{1}{c}{\multirow{2}{*}{\textbf{PCC ↑}}} &
  \multicolumn{1}{c}{\multirow{2}{*}{\textbf{SSIM ↑}}} &
  \multicolumn{1}{c}{\multirow{2}{*}{\textbf{MSE ↑}}} &
  \multicolumn{1}{c}{\multirow{2}{*}{\textbf{PCC ↑}}} &
  \multicolumn{1}{c}{\multirow{2}{*}{\textbf{SSIM ↑}}} \\
 &
  \multicolumn{1}{c}{} &
  \multicolumn{1}{c}{} &
  \multicolumn{1}{c}{} &
  \multicolumn{1}{c}{} &
  \multicolumn{1}{c}{} &
  \multicolumn{1}{c}{} &
  \multicolumn{1}{c}{} \\ \hline
\multirow{4}{*}{\textbf{S1}} &
  ShenDNN \citep{shen_deep_2019} &
  0.09$\pm$0.02 &
  0.23$\pm$0.23 &
  0.14$\pm$0.02 &
  69.10 &
  69.55 &
  \textbf{74.49} \\ \cline{2-8} 
 &
  ShenDNN+DGN \citep{shen_deep_2019} &
  \textbf{0.06$\pm$0.02} &
  \textbf{0.28$\pm$0.28} &
  0.46$\pm$0.06 &
  67.56 &
  66.79 &
  60.58 \\ \cline{2-8} 
 &
  ShenGAN \citep{shen_end--end_2019} &
  \textbf{0.06$\pm$0.02} &
  0.14$\pm$0.20 &
  0.48$\pm$0.13 &
  57.63 &
  66.60 &
  56.03 \\ \cline{2-8} 
 &
  BeliyEncDec \citep{beliy_voxels_2019} &
  \textbf{0.06$\pm$0.02} &
  0.16$\pm$0.21 &
  \textbf{0.60$\pm$0.08} &
  \textbf{72.82} &
  \textbf{76.35} &
  60.83 \\ \hline
\multirow{4}{*}{\textbf{S2}} &
  ShenDNN \citep{shen_deep_2019} &
  0.10$\pm$0.02 &
  0.24$\pm$0.21 &
  0.14$\pm$0.04 &
  72.31 &
  72.05 &
  \textbf{76.54} \\ \cline{2-8} 
 &
  ShenDNN+DGN \citep{shen_deep_2019} &
  0.07$\pm$0.01 &
  \textbf{0.25$\pm$0.26} &
  0.45$\pm$0.09 &
  64.62 &
  69.10 &
  56.67 \\ \cline{2-8} 
 &
  ShenGAN \citep{shen_end--end_2019} &
  \textbf{0.06$\pm$0.02} &
  0.15$\pm$0.20 &
  0.46$\pm$0.12 &
  58.40 &
  68.08 &
  53.46 \\ \cline{2-8} 
 &
  BeliyEncDec \citep{beliy_voxels_2019} &
  \textbf{0.06$\pm$0.02} &
  0.20$\pm$0.17 &
  \textbf{0.60$\pm$0.10} &
  \textbf{73.21} &
  \textbf{76.15} &
  58.01 \\ \hline
\multirow{4}{*}{\textbf{S3}} &
  ShenDNN \citep{shen_deep_2019} &
  0.10$\pm$0.02 &
  0.23$\pm$0.22 &
  0.13$\pm$0.02 &
  67.18 &
  68.85 &
  \textbf{73.08} \\ \cline{2-8} 
 &
  ShenDNN+DGN \citep{shen_deep_2019} &
  0.07$\pm$0.02 &
  \textbf{0.25$\pm$0.28} &
  0.45$\pm$0.05 &
  65.71 &
  64.23 &
  57.82 \\ \cline{2-8} 
 &
  ShenGAN \citep{shen_end--end_2019} &
  \textbf{0.06$\pm$0.02} &
  0.19$\pm$0.21 &
  0.50$\pm$0.13 &
  60.77 &
  72.44 &
  57.31 \\ \cline{2-8} 
 &
  BeliyEncDec \citep{beliy_voxels_2019} &
  \textbf{0.06$\pm$0.02} &
  0.21$\pm$0.19 &
  \textbf{0.60$\pm$0.07} &
  \textbf{69.04} &
  \textbf{78.01} &
  59.68 \\ \hline
\multirow{4}{*}{\textbf{Average result}} &
  ShenDNN \citep{shen_deep_2019} &
  0.10$\pm$0.00 &
  0.23$\pm$0.00 &
  0.14$\pm$0.00 &
  69.53$\pm$2.59 &
  70.15$\pm$1.68 &
  \textbf{74.70$\pm$1.74} \\ \cline{2-8} 
 &
  ShenDNN+DGN \citep{shen_deep_2019} &
  0.07$\pm$0.00 &
  \textbf{0.26$\pm$0.01} &
  0.45$\pm$0.01 &
  65.96$\pm$1.49 &
  66.71$\pm$2.44 &
  58.35$\pm$2.01 \\ \cline{2-8} 
 &
  ShenGAN \citep{shen_end--end_2019} &
  \textbf{0.06$\pm$0.00} &
  0.16$\pm$0.03 &
  0.48$\pm$0.02 &
  58.93$\pm$1.64 &
  69.04$\pm$3.03 &
  55.60$\pm$1.96 \\ \cline{2-8} 
 &
  BeliyEncDec \citep{beliy_voxels_2019} &
  \textbf{0.06$\pm$0.00} &
  0.19$\pm$0.02 &
  \textbf{0.60$\pm$0.00} &
  \textbf{71.69$\pm$2.30} &
  \textbf{76.84$\pm$1.02} &
  59.51$\pm$1.42 \\ \hline
\end{tabular}
}
%\arrayrulecolor{black}
\end{table*}

\section{Regions of interest in fMRI}
%The smallest unit of the three-dimensional fMRI images is a tiny cube called a voxel. Each voxel can capture the activity of a million brain cells by measuring the blood-oxygen-level-dependent signal \citep{ogawa_brain_1990}.
A common approach to fMRI analysis involves extracting signals from sections of the brain, called regions of interest (ROIs). %The advantage of an ROI-based analysis is the reduced number of voxels and, as a result, more sensitive analysis \citep{poldrack_region_2007}. 
The hierarchical nature of information processing in the visual cortex is shown in \citep{poldrack_progress_2015}. According to this finding, low-level image features (such as shapes and orientations) show a high correlation with the brain response signal in lower ROIs---the lower visual cortex (LVC). 
In contrast, high-level image features (such as semantic object categories) are shown to be encoded in higher ROIs, that is, the higher visual cortex (HVC). 
The lower visual cortex (LVC) consists of lower-order visual areas (V1, V2, and V3). Higher-order visual areas include the lateral occipital complex (LOC), \linebreak the fusiform face area (FFA), and the parahippocampal place area (PPA). Table \ref{table:ROIs} shows the details of the fMRI dataset's ROIs and the number of voxels per ROI in publicly available benchmark datasets.

\begin{table*}[]
\caption{Details of the fMRI dataset's ROIs and the number of voxels per ROI in publicly available benchmark datasets. The subject number is specified in brackets. n/a represents information not explicitly provided.}
\label{table:ROIs}
\resizebox{\textwidth}{!}{
\begin{tabular}{lllll}
\hline
\multicolumn{1}{c}{\textbf{Reference}}                & \multicolumn{2}{c}{\textbf{Dataset}}                                  & \multicolumn{1}{c}{\textbf{ROIs}} & \multicolumn{1}{c}{\textbf{Voxels}}                                                              \\ \hline
%van Gerven et al. \citep{van_gerven_neural_2010}       & \multicolumn{2}{l}{\texttt{69}}                         & V1, V2, V3,                        & 3092 (S1)                                                                                       \\ \hline
%Schoenmakers et al. \citep{schoenmakers_linear_2013}   & \multicolumn{2}{l}{\texttt{BRAINS}}                 & V1, V2                            & 2536 (S1), 2784 (S2), 2420 (S3)                                                                 \\ \hline
%Miyawaki et al. \citep{miyawaki_visual_2008}           & \multicolumn{2}{l}{\texttt{Binary contrast patterns}}                      & V1, V2                            & 797 (S1)                                                                                        \\ \hline
 \cite{kay_identifying_2008}                & \multicolumn{2}{l}{\texttt{vim-1}}                                                  & V1, V2, V3, V3A, V3B, V4, LO  & 25915 (S1), 26329 (S2)                                                                          \\ \hline
 \cite{vanrullen_reconstructing_2019} & \multicolumn{2}{l}{\texttt{Faces}}                     & n/a                               & n/a                                                                                             \\ \hline
 \cite{horikawa_generic_2017}     & \multicolumn{2}{l}{\texttt{Generic Object decoding }}                       & V1, V2, V3, V4, LOC, FFA, PPA     & \begin{tabular}[c]{@{}l@{}}4466 (S1), 4404(S2), 4643 (S3), \\ 4133 (S4), 4370 (S5)\end{tabular} \\ \hline
\multirow{3}{*}{ \cite{shen_deep_2019}}    & \multirow{3}{*}{\texttt{Deep Image Reconstruction}} & natural images       & V1, V2, V3, V4, LOC, FFA, PPA     & 11726 (S1), 11114 (S2), 9919 (S3)                                                               \\ \cline{3-5} 
                                                      &                                                     & artificial shapes    & V1, V2, V3, V4, LOC, FFA, PPA     & 11726 (S1), 11114 (S2), 9919 (S3)                                                               \\ \cline{3-5} 
                                                      &                                                     & alphabetical letters & V1, V2, V3, V4, LOC, FFA, PPA     & 11726 (S1), 11114 (S2), 9919 (S3)                                                               \\ \hline
\end{tabular}
}
\end{table*}

\bibliographystyle{frontiersinSCNS_ENG_HUMS} % for Science, Engineering and Humanities and Social Sciences articles, for Humanities and Social Sciences articles please include page numbers in the in-text citations
\bibliography{references}